\documentclass[journal]{IEEEtran}

\usepackage{graphicx}

\usepackage{amsmath}
\usepackage{amssymb}
\usepackage{amsfonts}

\usepackage{booktabs}
\usepackage{multirow}
\usepackage{subfig}
\usepackage{cite}
\usepackage{floatflt}
\usepackage{wrapfig}
\usepackage[linesnumbered, ruled,vlined]{algorithm2e}
\usepackage{moresize}
\usepackage{url}

\usepackage{tablefootnote}
\usepackage{bigfoot}

\usepackage{xcolor}
\usepackage{adjustbox}

\usepackage{colortbl}
\usepackage{tabularx}

\usepackage[whole]{bxcjkjatype}

\def\vector#1{\mbox{\boldmath $#1$}}

\newcommand{\argmin}{\mathop{\rm argmin}\limits}

\ifCLASSINFOpdf
\else
\fi
\begin{document}
%


\title{A Framework to Handle Multi-modal Multi-objective Optimization in Decomposition-based Evolutionary Algorithms}


%
%
%

\author{
Ryoji~Tanabe,~\IEEEmembership{Member,~IEEE,}and~Hisao~Ishibuchi,~\IEEEmembership{Fellow,~IEEE}
\thanks{R. Tanabe and H. Ishibuchi are with
Shenzhen Key Laboratory of Computational Intelligence, University Key Laboratory of Evolving Intelligent Systems of Guangdong Province, Department of Computer Science and Engineering, Southern University of Science and Technology, Shenzhen 518055, China. e-mail: (rt.ryoji.tanabe@gmail.com, hisao@sustc.edu.cn). (Corresponding author: Hisao Ishibuchi)}
}
\maketitle


\begin{abstract}

Multi-modal multi-objective optimization is to locate (almost) equivalent Pareto optimal solutions as many as possible.
While decomposition-based evolutionary algorithms have good performance for multi-objective optimization, they are likely to perform poorly for multi-modal multi-objective optimization due to the lack of mechanisms to maintain the solution space diversity.
To address this issue, this paper proposes a framework to improve the performance of decomposition-based evolutionary algorithms for multi-modal multi-objective optimization.
Our framework is based on three operations: assignment, deletion, and addition operations.
One or more individuals can be assigned to the same subproblem to handle multiple equivalent solutions.
In each iteration, a child is assigned to a subproblem based on its objective vector, i.e., its location in the objective space.
The child is compared with its neighbors in the solution space assigned to the same subproblem.
The performance of improved versions of six decomposition-based evolutionary algorithms by our framework is evaluated on various test problems regarding the number of objectives, decision variables, and equivalent Pareto optimal solution sets.
Results show that the improved versions perform clearly better than their original algorithms.

\end{abstract}


%
\IEEEpeerreviewmaketitle

\section{Introduction}
\label{sec:introduction}

\IEEEPARstart{M}ULTI-OBJECTIVE optimization problems (MOPs) appear in real-world applications.
Since no solution $\vector{x}$ can simultaneously minimize multiple objective functions in general, the goal of MOPs is usually to find a Pareto optimal solution preferred by a decision maker \cite{Miettinen98}.
When the decision maker's preference information is unavailable a priori, an ``a posteriori'' decision making is conducted.
The decision maker selects the final solution $\vector{x}^{\rm final}$ from a set of solutions that approximates the Pareto front in the objective space.



An evolutionary multi-objective optimization algorithm (EMOA) is frequently used for the ``a posteriori'' decision making.
Since EMOAs are population-based optimizers, they are likely to find a set of solutions in a single run.
A number of EMOAs have been proposed in the literature.
They can be classified into the following three categories: dominance-based EMOAs (e.g., NSGA-II \cite{DebAPM02}), indicator-based EMOAs (e.g., IBEA \cite{ZitzlerK04}), and decomposition-based EMOAs (e.g., MOEA/D \cite{ZhangL07} and NSGA-III \cite{DebJ14}).
In particular, decomposition-based EMOAs have shown promising performance \cite{TrivediSSG17}.

\begin{figure}[t]
  \newcommand{\widthvar}{0.38}
  \begin{center} 
\includegraphics[width=\widthvar\textwidth]{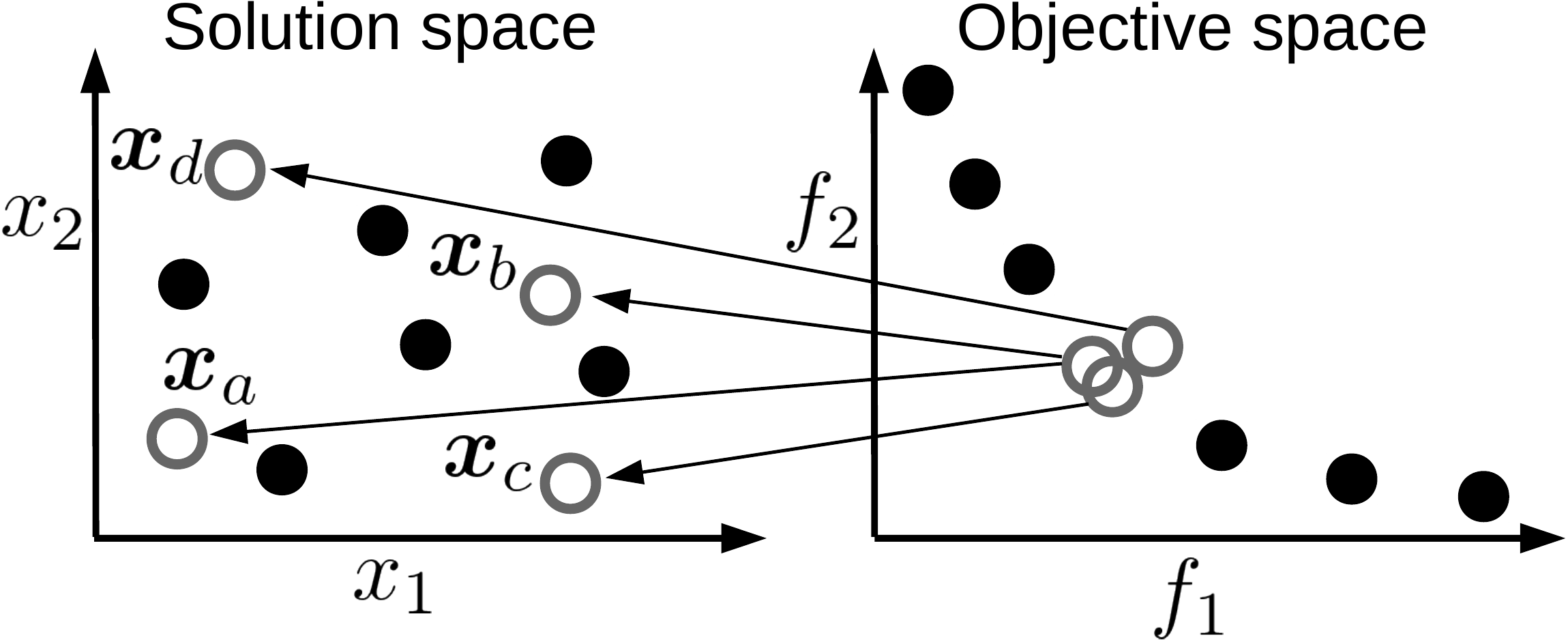}
\caption{
\small
Illustration of a situation where three solutions are almost the same in the objective space but dissimilar in the solution space.
%
}
\label{fig:mmop_example}
  \end{center}
\end{figure}


In the EMO community, it has been implicitly assumed that the decision maker is interested only in the distribution of solutions in the objective space.
Thus, the distribution of solutions in the solution space has not received much attention.
However, after the decision maker has selected the final solution $\vector{x}^{\rm final}$ based on its objective vector $\vector{f}(\vector{x}^{\rm final})$, she/he may want to examine other dissimilar solutions with equivalent quality or slightly inferior quality  \cite{SebagTTLB05,RudolphP09,SchutzeVC11}.
%

Fig. \ref{fig:mmop_example} shows a situation where the four solutions $\vector{x}_a$, $\vector{x}_b$, $\vector{x}_c$, and $\vector{x}_d$ are far from each other in the solution space but close to each other in the objective space.
Although $\vector{x}_d$ is a dominated solution, it may be acceptable for the decision maker.
This is because the difference between $\vector{f}(\vector{x}_a)$, $\vector{f}(\vector{x}_b)$, $\vector{f}(\vector{x}_c)$, and $\vector{f}(\vector{x}_d)$ is small enough.
If the decision maker has obtained multiple dissimilar solutions with similar quality, she/he can select $\vector{x}^{\rm final}$ according to her/his preference in the solution space.
For example, suppose that $\vector{x}_a$ in Fig. \ref{fig:mmop_example} becomes unavailable due to some accident (e.g., material shortages and mechanical failures) after the decision maker has selected $\vector{x}_a$ as $\vector{x}^{\rm final}$.
In such a case, she/he can substitute one of $\vector{x}_b$, $\vector{x}_c$, and $\vector{x}_d$ for $\vector{x}_a$.
Sch{\"{u}}tze et al. give a more practical example on space mission design problems \cite{SchutzeVC11}.
Sebag et al. also demonstrate the importance of multiple equivalent solutions for the decision maker on functional brain imaging problems \cite{SebagTTLB05}.
Previous studies on other real-world problems with multiple equivalent solutions include diesel engine design problems \cite{HiroyasuNM05}, distillation plant layout problems \cite{PreussKBH08}, and rocket engine design problems \cite{KudoYF11}.

A multi-modal multi-objective problem (MMOP) is to locate (almost) equivalent Pareto optimal solutions as many as possible \cite{SebagTTLB05,DebT08}.
On the one hand, it is sufficient for MOPs to find one of $\vector{x}_a$, $\vector{x}_b$, and $\vector{x}_c$ in Fig. \ref{fig:mmop_example} since their quality is almost the same.
On the other hand, all of $\vector{x}_a$, $\vector{x}_b$, $\vector{x}_c$, and $\vector{x}_d$ should be found for MMOPs.
Since multiple equivalent solutions play crucial roles in the reliable decision making process as explained above, MMOPs are important in practice.
Evolutionary multi-modal multi-objective optimization algorithms (EMMAs) are specially designed optimizers for MMOPs.
Unlike EMOAs, EMMAs have mechanisms to handle multiple equivalent solutions.
Representative EMMAs include $P_{Q, \epsilon}$-MOEA \cite{SchutzeVC11}, Omni-optimizer \cite{DebT08}, DIOP \cite{UlrichBT10}, Niching-CMA \cite{ShirPNE09}, and MO\_Ring\_PSO\_SCD \cite{YueQL17}.





In our earlier study \cite{TanabeI18}, we proposed MOEA/D-AD, which is a specially designed MOEA/D for MMOPs.
MOEA/D-AD can assign multiple individuals to each subproblem in order to handle equivalent solutions.
For each iteration, a child $\vector{u}$ is assigned to a subproblem whose weight vector is closest to its objective vector $\vector{f}(\vector{u})$, in terms of the perpendicular distance.
Then, $\vector{u}$ is compared to the individuals assigned to the same subproblem based on the weighted Tchebycheff function and the contribution to the solution space diversity.



In this paper, we extend our earlier study \cite{TanabeI18} and propose a general framework (called ADA) to handle multi-modal multi-objective optimization in a variety of decomposition-based EMOAs (including reference-based algorithms such NSGA-III \cite{DebJ14} and $\theta$-DEA \cite{YuanXWY16}).
The proposed ADA framework uses assignment, deletion, and addition operations.
While decomposition-based EMOAs work well for multi-objective optimization, they are likely to perform poorly for multi-modal multi-objective optimization.
This is because most of them do not have mechanisms to maintain the solution space diversity.
In this paper, we show that efficient EMMAs could be realized by incorporating a solution space diversity maintenance mechanism into decomposition-based EMOAs.
The proposed ADA framework facilitates the solution space diversity maintenance of existing EMOAs.
We incorporate ADA into the following six representative decomposition-based EMOAs: MOEA/D-AGR \cite{WangZZGJ16}, MOEA/D-DU \cite{YuanXWZY16}, eMOEA/D \cite{JiangYWL18}, NSGA-III \cite{DebJ14}, $\theta$-DEA \cite{YuanXWY16}, and RVEA \cite{ChenJOS16}.
We examine the ability of the ADA versions of those EMOAs (MOEA/D-AGR-ADA, MOEA/D-DU-ADA, eMOEA/D-ADA, NSGA-III-ADA, $\theta$-DEA-ADA, and RVEA-ADA) to locate multiple equivalent Pareto optimal solutions on various test problems.
We also analyze the behavior of those ADA versions.

The differences between this paper and our previous study \cite{TanabeI18} are as follows:

\begin{enumerate}
\item We propose ADA. While MOEA/D-AD \cite{TanabeI18} is a single algorithm for MMOPs, ADA  is a framework to improve the performance of the existing decomposition-based EMOAs for MMOPs.
In contrast to MOEA/D-AD, each configuration (e.g., the assignment operation) in ADA depends on an EMOA to be combined.
\item We demonstrate that ADA can be combined with the six decomposition-based EMOAs, including so-called reference vector-based EMOAs. MOEA/D-AD does not have such flexibility.
\item We introduce a practical two-phase decision making method with a solution set found by an ADA-based algorithm. We also introduce two post-processing methods for benchmarking an ADA-based algorithm.
\item While we used only two-objective and two-variable problems in \cite{TanabeI18}, we use problems with a various number of objectives $M$, design variables $D$, and equivalent Pareto optimal solution subsets $O$ to investigate the scalability of EMMAs. Such a scale-up study has not been performed in the literature. We also use problems with distance-related variables \cite{ZhangSA17,LiuYG18}.
\item We compare the six ADA-based algorithms to three state-of-the-art EMMAs, including TriMOEA-TA\&R \cite{LiuYG18} proposed after the publication of \cite{TanabeI18}.  
\end{enumerate}

The rest of this paper is organized as follows.
Section \ref{sec:preliminaries} provides some preliminaries of this paper.
Section \ref{sec:proposed_method} introduces ADA and explains how to incorporate it into decomposition-based EMOAs.
Section \ref{sec:experimental_settings} describes the experimental setup. 
Section \ref{sec:experimental_results} examines the performance of the six ADA variants.
Section \ref{sec:conclusion} concludes this paper. 


\newtheorem{defi}{Definition}

\section{Preliminaries}
\label{sec:preliminaries}


First, Subsections \ref{sec:def_MOPs} and \ref{sec:def_MMOPs} give definitions of MOPs and MMOPs, respectively.
Then, Subsections \ref{sec:dmoea} and \ref{sec:rmoea} describe decomposition-based EMOAs and reference vector-based EMOAs, respectively.
Subsections \ref{sec:dmoea} and \ref{sec:rmoea} mainly explain components of the six existing EMOAs used in ADA.
Algorithms S.1--S.6 in the supplementary file provide the overall procedures of the six EMOAs.
Section \ref{sec:proposed_method} introduces how each component is used in ADA.





\subsection{Definition of MOPs}
\label{sec:def_MOPs}

A continuous MOP is to find a solution $\vector{x} \in \mathbb{S} \subseteq \mathbb{R}^D$ that minimizes a given objective function vector $\vector{f}: \mathbb{S} \rightarrow \mathbb{R}^M, \vector{x} \mapsto \vector{f}(\vector{x})$.
$\mathbb{S} = \prod^D_{j=1} [x^{\rm min}_j, x^{\rm max}_j]$ is the $D$-dimensional solution space where $x^{\rm min}_j \leq x_j \leq x^{\rm max}_j$ for each index $j \in \{1, ..., D\}$.
$\mathbb{R}^M$ is the $M$-dimensional objective space.


A solution $\vector{x}_1$ is said to dominate $\vector{x}_2$ if and only if $f_i (\vector{x}_1) \leq f_i (\vector{x}_2)$ for all $i \in \{1, ..., M\}$ and $f_i (\vector{x}_1) < f_i (\vector{x}_2)$ for at least one index $i$.
If $ \vector{x}^*$ is not dominated by any other solutions, it is called a Pareto optimal solution.
The set of all $\vector{x}^*$ is the Pareto optimal solution set, and the set of all $\vector{f}(\vector{x}^*)$ is the Pareto front.
The goal of MOPs for the ``a posteriori'' decision making is to find a non-dominated solution set that approximates the Pareto front in the objective space.





\subsection{Definition of MMOPs}
\label{sec:def_MMOPs}


Although the term ``multi-modal multi-objective optimization'' was firstly coined in \cite{DebT05,SebagTTLB05} in 2005, the definition of MMOPs has not been explicitly given in most previous studies.
We consider the following two types of MMOPs:
(i) {\em Type1-MMOP} is to locate all Pareto optimal solutions.
(ii) {\em Type2-MMOP} is to locate all Pareto optimal solutions and non-Pareto optimal solutions which have acceptable quality for the decision maker. Those non-Pareto optimal solutions should be far from the Pareto optimal solutions and the other non-Pareto optimal solutions in the solution space.



For example, $\vector{x}_a$, $\vector{x}_b$, and $\vector{x}_c$ in Fig. \ref{fig:mmop_example} should be found for Type1-MMOPs.
In addition, the non-Pareto optimal solution $\vector{x}_d$ should be found for Type2-MMOPs if its quality $\vector{f}(\vector{x}_d)$ is acceptable to the decision maker.
While most existing studies (e.g., \cite{DebT08,YueQL17}) assume Type1-MMOPs, only a few studies (e.g., \cite{SebagTTLB05,SchutzeVC11,UlrichBT10}) address Type2-MMOPs.
Although there is room for discussion, Type2-MMOPs may be more practical than Type1-MMOPs.
Diverse solutions with similar quality to the final solution $\vector{x}^{\rm final}$ are beneficial to the decision maker even if their quality is slightly worse than $\vector{x}^{\rm final}$ \cite{SebagTTLB05,SchutzeVC11}.

Nevertheless, we focus only on Type1-MMOPs in this paper.
The main reason is due to the difficulty in benchmarking EMMAs on Type2-MMOPs.
In contrast to Type1-MMOPs, Type2-MMOPs are loosely defined.
This is because the terms  ``acceptable quality'' and ``far from'' in Type2-MMOPs significantly depend on the decision maker.
It is difficult to define the two key factors in Type2-MMOPs for benchmarking purposes in a fair manner.
How to evaluate the performance of EMMAs for Type2-MMOPs itself is another research topic.


\subsection{Decomposition-based EMOAs (MOEA/D-type algorithms)}
\label{sec:dmoea}


Although some frameworks of decomposition-based EMOAs have been proposed in the literature, MOEA/D-type algorithms show the promising performance \cite{TrivediSSG17}.
Here, we describe MOEA/D-type algorithms.
First, we explain the most basic MOEA/D \cite{ZhangL07} and its modified version called MOEA/D-DE \cite{LiZ09}.
The framework of MOEA/D-DE is used in most MOEA/D-type algorithms.
Then, we describe components of three MOEA/D-type algorithms: MOEA/D-AGR \cite{WangZZGJ16}, MOEA/D-DU \cite{YuanXWZY16}, and eMOEA/D \cite{JiangYWL18}.





\subsubsection{MOEA/D}
\label{sec:moead}

The most basic MOEA/D \cite{ZhangL07} decomposes an $M$-objective MOP into $N$ single-objective subproblems using a scalarizing function $g: \mathbb{R}^M \rightarrow \mathbb{R}$ and a set of uniformly distributed weight vectors $\vector{W} = \{\vector{w}_1, ..., \vector{w}_{N}\}$.
For each $i \in \{1, ..., N\}$, $\vector{w}_i = (w_{i,1}, ..., w_{i,M})^{\rm T}$, and $\sum^M_{k=1} w_{i,k} = 1$.
The $j$-th individual $\vector{x}_j$ in the population $\vector{P}$ is assigned to the $j$-th subproblem with $\vector{w}_j$.
Thus, the population size $\mu$ is always equal to the number of weight vectors $N$.
The $j$-th subproblem also has its neighborhood index list $\vector{B}_{j} = \{b_{j,1}, ...,b_{j,S}\}$, which consists of indices of the $S$ closest weight vectors to $\vector{w}_j$ in the weight vector space.


After the initialization of $\vector{P}$ and $\vector{W}$, the following steps are repeatedly performed until a termination condition is satisfied.
In each iteration $t$, for the $j$-th subproblem ($j \in \{1, ..., N\}$), an index list $\vector{T}$ is set to $\vector{B}_j$.
Two indices $a$ and $b$ of the parent individuals $\vector{x}_a$ and $\vector{x}_b$ are randomly selected from $\vector{T}$.
$\vector{u}$ is generated by applying variation operators to $\vector{x}_a$ and $\vector{x}_b$.
The SBX crossover and the polynomial mutation \cite{DebA95} are used in the original MOEA/D.
After $\vector{u}$ has been generated, the replacement selection is applied to each neighborhood subproblem $k \in \vector{T}$.
The current solution $\vector{x}_k$ of the $k$-th subproblem is replaced with $\vector{u}$ if $g(\vector{u}|\vector{w}_k) \leq g(\vector{x}_k|\vector{w}_k)$.


\subsubsection{MOEA/D-DE}
\label{sec:moeadde}

The differences between MOEA/D and MOEA/D-DE are threefold.
First, the differential evolution (DE) operator \cite{StornP97} is used instead of the SBX crossover \cite{DebA95}.
Second, two types of index lists ($\vector{B}_{j}$ and $\{1, ..., N\}$) are used.
The index list $\vector{T}$ is set to $\vector{B}_{j}$ with a probability of $\delta \in [0,1]$ and $\{1, ..., N\}$ with a probability of $1-\delta$.
Thus, $\vector{T}$ can be set to all individual indices.
Third, the maximum number of individuals replaced by the child $\vector{u}$ is restricted to $n^{\rm rep}$.




\subsubsection{MOEA/D-AGR}
\label{sec:moeadagr}





The difference between MOEA/D-AGR and MOEA/D-DE is caused by a replacement method.
In MOEA/D-AGR, indices of subproblems to be updated by a newly generated solution $\vector{u}$ are set to $K$ neighborhood indices of the $j$-th subproblem in the weight vector space where $j$ is the subproblem index whose scalarizing function value $g(\vector{u}|\vector{w}_j)$ is the best among all $N$ subproblems:
\begin{align}
\label{eqn:moeadagr_selection}
j =  \argmin_{k \in \{1, ..., N\}} \bigl\{g(\vector{u}|\vector{w}_k)\bigr\}.
\end{align}

The weighted Tchebycheff function $g^{\rm tch}$ is used in \eqref{eqn:moeadagr_selection}:
\begin{align}
\label{eqn:tchebycheff-mul}
g^{\rm tch}(\vector{x} | \vector{w}) = \max_{i \in \{1, ..., M\}} \bigl\{ w_i |f_i (\vector{x}) - z^*_i|  \bigr\},
\end{align}
where $\vector{z}^* = (z^*_1, ..., $ $z^*_M)^{\rm T}$ is the ideal point.
Since finding the true ideal point is difficult in general, its approximation is used in \eqref{eqn:tchebycheff-mul}.
The $i$-th element of the approximation of $\vector{z}^*$ is the minimum objective value found during the search process.

The replacement neighborhood size $K$ plays a crucial role in balancing exploration and exploitation in MOEA/D-AGR in a similar manner to $S$ in MOEA/D.
A large $K$ value encourages exploitation.
In MOEA/D-AGR, the $K$ value deterministically increases with the number of iterations.
The results presented in \cite{WangZZGJ16} show that a scheduling method based on the sigmoid function is suitable for MOEA/D-AGR.



\subsubsection{MOEA/D-DU}
\label{sec:moeaddu}

%


In contrast to MOEA/D-AGR, the selection of subproblems that need to be updated is based on the distance in the objective and weight vector spaces in MOEA/D-DU.
In the normalized objective space, the perpendicular distance between the normalized objective vector $\vector{f}'(\vector{u})$ and $\vector{w}_j$ is calculated for each $j \in \{1, ..., N\}$.
The replacement is performed for $K$ subproblems with the minimum perpendicular distance.
Similar to MOEA/D-AGR, $K$ controls the balance between exploration and exploitation.


MOEA/D-DU uses the division version of the weighted Tchebycheff function $g^{\rm dtch}$ \cite{QiMLJSW14}:
%
\begin{align}
\label{eqn:tchebycheff-div}
g^{\rm dtch}(\vector{x} | \vector{w}) = \max_{i \in \{1, ..., M\}} \left\{ \frac{|f_i (\vector{x}) - z^*_i|}{w_i}\right\},
\end{align}
if $w_i = 0$, it is set to $10^{-6}$ to avoid division by zero.
It is reported in \cite{QiMLJSW14} that the distribution of the search directions of MOEA/D with $g^{\rm dtch}$ is more uniform than that with $g^{\rm tch}$.

\subsubsection{eMOEA/D}
\label{sec:emoead}


In eMOEA/D, the replacement method is applied to $K$ subproblems with the minimum scalarizing function values.
MOEA/D-AGR and eMOEA/D are the same regarding the use of scalarizing function values to select subproblems.
The following multiplicative scalarizing function (MSF) \cite{JiangYWL18} is used in eMOEA/D:
%
\begin{align}
\label{eqn:msf}
g^{\rm msf}(\vector{x} | \vector{w}) =\frac{\Bigl(\max_{i \in \{1, ..., M\}}\bigl\{\frac{1}{w_i}|f_i(\vector{x} - z^*_i)|\bigr\}\Bigr)^{1+\alpha}}{\Bigl(\min_{i \in \{1, ..., M\}}\bigl\{\frac{1}{w_i}|f_i(\vector{x} - z^*_i)|\bigr\}\Bigr)^{\alpha}},
\end{align}
%
where $\alpha$ controls the size of the so-called improvement region.
$\alpha$ plays a similar role in the penalty value $\theta$ of the PBI function in \eqref{eqn:pbi}, which will be explained later.
Even if $\vector{f}(\vector{x})$ is far from $\vector{w}_j$ regarding the the perpendicular distance, $\vector{x}$ would be evaluated as being superior using a sufficiently large $\alpha$ value.
In \eqref{eqn:msf}, $g^{\rm msf}$ with $\alpha = 0$ is identical to $g^{\rm dtch}$.


According to a general rule of thumb ``emphasize diversity and convergence at the early and later stages, respectively'', $\alpha$ decreases linearly with the number of iterations $t$:
\begin{align}
\label{eqn:alpha}
\alpha = \beta \biggl(1 - \frac{t}{t^{\rm max}}\biggr) \biggl(M \Bigl(\min_{k \in \{1, ..., M\}} \{w_k\}\Bigr)\biggr),
\end{align}
where $t^{\rm max}$ is the maximum number of iterations.
The recommended value of $\beta$ is $1$.


\subsection{Reference vector-based EMOAs}
\label{sec:rmoea}


Representative reference vector-based EMOAs include NSGA-III \cite{DebJ14}, $\theta$-DEA \cite{YuanXWY16}, RVEA \cite{ChenJOS16}, VaEA \cite{XiangZLC16}, and SPEA/R \cite{JiangY17a}.
While $\vector{w} \in \vector{W}$ is called the weight vector in decomposition-based EMOAs, it is referred to the reference vector in reference vector-based EMOAs.
Below, we explain NSGA-III, $\theta$-DEA, and RVEA.



\subsubsection{NSGA-III}

NSGA-III is an improved version of NSGA-II \cite{DebAPM02} for many-objective optimization.
For each iteration $t$, children $\vector{Q}$ are generated by applying variation operators to randomly selected pairs of individuals from $\vector{P}$.
In the environmental selection, $\mu$ individuals for the next iteration $t+1$ are selected from the union of $\vector{P}$ and $\vector{Q}$.
The primary and secondary criteria are based on the non-domination levels and the reference vector-based niching method, respectively.

Individuals in the union $\vector{P} \cup \vector{Q}$ are grouped as $\vector{F}_1, \vector{F}_2, ...$ according to their non-domination levels.
First, $\vector{P}$ and the front index $i$ are initialized as $\vector{P}=\emptyset$ and $i=1$, respectively.
Then, individuals in $\vector{F}_i$ are added to $\vector{P}$ and $i$ is incremented until $|\vector{P}|+ |\vector{F}_i|\geq \mu$.
After this operation, $\vector{P} = \vector{F}_1  \cup  ... \cup \vector{F}_{l-1}$, where $l$ is the index of the last front.
If $|\vector{P}| < \mu$, other $\mu - |\vector{P}|$ individuals are selected from the last front $\vector{F}_l$ using the niching method described below.
At the beginning of the niching procedure, an individual $\vector{x}$ in $\vector{P}$ and $\vector{F}_l$ is assigned to the $j$-th subproblem with the minimum perpendicular distance between its normalized objective vector $\vector{f}'(\vector{x})$ and   $\vector{w}_j$:
\begin{align}
\label{eqn:nsgaiii_selection}
j =  \argmin_{k \in \{1, ..., N\}} \Bigl\{{\rm PD}\bigl(\vector{f}'(\vector{x}), \vector{w}_k\bigr)\Bigr\},
\end{align}
where the function ${\rm PD}$ returns the perpendicular distance between two input vectors.
After the assignments of all individuals, other individuals in the next iteration are selected from $\vector{F}_l$ based on the number of individuals assigned to each subproblem and their perpendicular distance.

\subsubsection{$\theta$-DEA}

The environmental selection in $\theta$-DEA is similar to that of NSGA-III.
However, the secondary criterion in $\theta$-DEA is based on the so-called $\theta$-dominance.
First, each individual in the union of $\vector{P}$ and $\vector{F}_l$ is assigned to the $j$-th subproblem using \eqref{eqn:nsgaiii_selection}.
Then, individuals assigned to each subproblem are ranked based on their $\theta$-dominance levels.
Let us assume that two individuals $\vector{x}$ and $\vector{y}$ are assigned to the same $j$-th subproblem.
$\vector{x}$ is said to $\theta$-dominate $\vector{y}$ if $g^{\rm pbi}(\vector{x}|\vector{w}_j) < g^{\rm pbi}(\vector{y}|\vector{w}_j)$.
The PBI function $g^{\rm pbi}$ is given as:
\begin{align}
\label{eqn:pbi}
g^{\rm pbi}(\vector{x} | \vector{w}) &= d_1 + \theta \, d_2,\\
\label{eqn:pbi_d1}
d_1 &= \frac{\| \left(\vector{f} (\vector{x}) - \vector{z}^* \right)^{\rm T} \, \vector{w}\|}{\|\vector{w}\|},\\
\label{eqn:pbi_d2}
d_2 &= \left\| \vector{f} (\vector{x}) - \left(\vector{z}^* +  d_1 \, \frac{\vector{w}}{\|\vector{w}\|} \right)\right\|,
\end{align}
where $\|\vector{a}\|$ indicates the Euclidean norm of $\vector{a}$.
The distance $d_1$ represents how close the objective vector $\vector{f} (\vector{x})$ is to the Pareto front, and 
$d_2$ is the perpendicular distance between $\vector{f} (\vector{x})$ and $\vector{w}$.
The penalty parameter $\theta$ balances the convergence ($d_1$) and the diversity ($d_2$).
The recommended $\theta$ value is $10^6$ for $M$ vectors $\vector{w}$ with the axis directions (e.g., $(1, 0, ..., 0)^{\rm T}$) and  $5$ for all the other vectors $\vector{w}$.




\subsubsection{RVEA}


RVEA uses a set of unit reference vectors $\vector{V} = \{\vector{v}_1, ..., \vector{v}_N\}$, instead of a set of reference vectors $\vector{W} = \{\vector{w}_1, ..., \vector{w}_N\}$, where $\vector{v}_i = \vector{w}_i/\|\vector{w}_i\|$ for each $i \in \{1, ..., N\}$.
RVEA adaptively adjusts $\vector{V}$ based on the current $\vector{P}$.



After children $\vector{Q}$ have been generated, the environmental selection is applied to the union of $\vector{P}$ and $\vector{Q}$.
First, for each individual in $\vector{P} \cup \vector{Q}$, $\vector{f}(\vector{x})$ is transformed as $\vector{f}'(\vector{x}) = \vector{f}(\vector{x}) - \vector{z}^*$.
Then, $\vector{x}$ in $\vector{P} \cup \vector{Q}$ is assigned to the $j$-th subproblem with the minimum angle between $\vector{f}'(\vector{x})$ and $\vector{v}_j$:
%
%
\begin{align}
  \label{eqn:rvea_selection}
j =  \argmin_{k \in \{1, ..., N\}} \Bigl\{{\rm angle}\bigl(\vector{f}'(\vector{x}), \vector{v}_k \bigl)\Bigr\},
\end{align}
where the function ${\rm angle}(\vector{a}, \vector{b})$ in \eqref{eqn:rvea_selection} returns the angle between the two input vectors $\vector{a}$ and $\vector{b}$.



Then, individuals assigned to the same subproblem are compared based on their angle-penalized distance (APD) values.
For each $j \in \{1, ..., N\}$, the best individual with the minimum APD value can survive to the next iteration.
The APD value of $\vector{x}$ with the unit reference vector $\vector{v}$ is given as:
\begin{align}
\label{eqn:apd}
  {\rm APD}(\vector{x}) &= \bigl(1+ P (\vector{x}, \vector{v} )\bigr) \; \|\vector{f}'(\vector{x})\|,\\
  P(\vector{x}, \vector{v}) &= M \, \Biggl( \frac{t}{t^{\rm max}}\Biggr)^{\alpha} \, \Biggl(\frac{{\rm angle}\bigl(\vector{f}'(\vector{x}), \vector{v}\bigr)}{\gamma(\vector{v})}\Biggr),\\
  \gamma(\vector{v}) &= \min_{\vector{s} \in \vector{V}\backslash\{\vector{v}\}} \bigr\{ {\rm angle}(\vector{v}, \vector{s}) \bigl\},
\end{align}
where $P(\vector{x}, \vector{v})$ is a penalty value for $\vector{x}$.
The larger the angle between $\vector{f}'(\vector{x})$ and $\vector{v}$ is, the larger the penalty value is given to $\vector{x}$.
$\gamma(\vector{v})$ is used to normalize the angle.
$t$ is the current number of iterations, and $t^{\rm max}$ is the maximum number of iterations.
The influence of the angle-based penalty scheme increases as the search progresses.
The recommended setting of $\alpha$ is $2$.





%






\section{Proposed ADA framework}
\label{sec:proposed_method}





This section explains the proposed ADA framework and the six ADA versions.
Unlike EMOAs for MOPs, EMMAs for MMOPs need to maintain the diversity of the population in both the objective and solution spaces.
For example, Omni-optimizer \cite{DebT08} uses an aggregate crowding distance metric in the objective and solution spaces.
While most EMMAs (e.g., \cite{ShirPNE09,UlrichBT10,YueQL17}) aggregate the objective and solution space diversity metrics similar to Omni-optimizer, ADA uses them in a two-phase manner similar to TriMOEA-TA\&R \cite{LiuYG18}.
More specifically, ADA handles the objective space diversity by an assignment method in an original EMOA and the solution space diversity by a simple niching criterion.

On the one hand, a single individual is assigned to each subproblem in most decomposition-based EMOAs.
Thus, the population size $\mu$ is equal to the number of weight/reference vectors $N$ (i.e., $\mu = N$).
On the other hand, one or more individuals can be assigned to each subproblem in the ADA framework (i.e., $\mu \geq N$).
This mechanism is to maintain multiple equivalent individuals in each subproblem.
The $\mu$ value is adaptively adjusted in ADA.





Algorithm \ref{alg:ada} shows the ADA framework.
Algorithms S.7--S.12 in the supplementary file show the six ADA-based algorithms.
Algorithms S.7--S.12 are almost the same to Algorithm \ref{alg:ada}.
Lines 3, 11, and 16 are different between Algorithms S.7--S.12 and Algorithm \ref{alg:ada}.
At the beginning of the search, $\mu$ is set to $N$ (line 1).
The population $\vector{P}$ and the weight/reference vector set $\vector{W}$ are also initialized.
The $i$-th individual $\vector{x}_i$ in $\vector{P}$ is assigned to the $i$-th subproblem (lines 2--3).
This operation is unnecessary for those EMOAs with a reassignment procedure of individuals to subproblems such as RVEA.
After the child $\vector{u}$ has been generated by the reproduction operations (line 6--8), the population $\vector{P}$ is updated using the assignment (line 10), deletion (lines 16--17), and addition (lines 18--19) operations.
The assignment and deletion operations differ depending on an EMOA to be combined with the ADA framework.
After the normalization of the $\mu + 1$ objective vectors (line 9), $\vector{u}$ is assigned to the $j$-th subproblem (line 10).
$\vector{X}$ denotes a set of individuals that have been assigned to the $j$-th subproblem and are in the neighborhood of $\vector{u}$ in the solution space (line 11).
Thus, ADA requires a neighborhood criterion in the solution space.
$\vector{X}$ is explained later in detail using Fig. \ref{fig:example_X}.
Two Boolean variables $b^{\rm winner}$ and $b^{\rm explorer}$ (line 12) are used in the addition operation (lines 18--19).
$\vector{u}$ enters $\vector{P}$ if it satisfies either of the following two conditions.
One is that there is no individual in $\vector{X}$ (lines 13--14).
The other is that $\vector{u}$ is better than at least one individual in $\vector{X}$ in the deletion operation (lines 15--17).




The following Subsections (from \ref{sec:ada_reprodction} to \ref{sec:ada_addition}) explain each step of ADA in detail.
Subsection \ref{sec:decision_making} presents an effective decision making based on a solution set found by the proposed approach.
Apart from the decision making, Subsection \ref{sec:ada_postprocessing} introduces two post-processing methods for benchmarking.
Subsection \ref{sec:ada_applicability} discusses the applicability of ADA.
Subsection \ref{sec:ada_originality} discusses the originality of ADA.

\subsection{Reproduction operation}
\label{sec:ada_reprodction}


ADA uses the basic GA operators (i.e., the SBX crossover and the polynomial mutation \cite{DebA95}) and the simplest method of selecting parents.
First, two parents  $\vector{x}_a$ and $\vector{x}_b$ are randomly selected from $\vector{P}$ such that $a \neq b$, regardless of the subproblem to which each individual has been assigned.
Then, $\vector{u}$ is reproduced by applying SBX to $\vector{x}_a$ and $\vector{x}_b$.
Finally, the polynomial mutation is applied to $\vector{u}$.
All ADA-based algorithms use the same mating selection scheme, regardless of the mating selection schemes in their original EMOAs.

%
%


In principle, any variation operators can be incorporated into ADA, including DE operators \cite{LiZ09}, PSO operators \cite{YueQL17}, and model-based methods \cite{ShirPNE09,ZhouZJ09}.
Any parent selection methods can also be used in ADA, including the distance-based selection methods in the solution space \cite{CastilloSAML17}.
The performance of ADA can be improved by using these more sophisticated methods.
However, if such effective methods are used in ADA, it is unclear which algorithmic component mainly contributes
to the overall performance of ADA-based algorithms.
Since we want to investigate the effectiveness of the ADA framework in an isolated manner, we use the simplest variation operators and parent selection method in this paper.

\def\HiLi{\leavevmode\rlap{\hbox to \hsize{\color{black!15}\leaders\hrule height .8\baselineskip depth .5ex\hfill}}}

\IncMargin{0.5em}
\begin{algorithm}[t]
\small
\SetSideCommentRight
$t \leftarrow 1$, $\mu \leftarrow N$, initialize the population $\vector{P} =\{ \vector{x}_{1}, ..., \vector{x}_{\mu}\}$ and the weight/reference vector set $\vector{W} =\{ \vector{w}_{1}, ..., \vector{w}_{N}\}$\;
\For{$i \in \{1, ..., N\}$}{
  Assign $\vector{x}_i$ to the $i$-th subproblem\;
}
\While{$\textsf{\upshape{The termination criteria are not met}}$}{
  $\mu \leftarrow |\vector{P}|$\;
Randomly select $a$ and $b$ from $\{1, ..., \mu\}$ such that $a \neq b$\;
Generate the child $\vector{u}$ by applying the crossover operation to $\vector{x}_a$ and $\vector{x}_b$\;
Apply the mutation operation to $\vector{u}$\;
%
Normalize the objective vectors $\vector{f}(\vector{x}_1), ..., \vector{f}(\vector{x}_{\mu}), \vector{f}(\vector{u})$\;
Assign $\vector{u}$ to the $j$-th subproblem\;
$\vector{X} \leftarrow \{\vector{x} \in \vector{P} | \vector{x}$ has been assigned to the $j$-th subproblem and is in the neighborhood of $\vector{u}$ in the solution space$\}$\;
$b^{\rm explorer} \leftarrow {\rm FALSE}$ and $b^{\rm winner} \leftarrow {\rm FALSE}$\;
\If{$\vector{X} = \emptyset$}{ 
  $b^{\rm explorer} \leftarrow {\rm TRUE}$\;
}
\For{$\vector{x} \in \vector{X}$}{
        \If{$\vector{x}$ $\textsf{\upshape{is worse than}}$ $\vector{u}$}{         
          $\vector{P} \leftarrow \vector{P} \backslash \{\vector{x}\}$ and $b^{\rm winner} \leftarrow {\rm TRUE}$
        }
}
\If{$b^{\rm winner} = {\rm TRUE}$ $\textsf{\upshape{or}}$ $b^{\rm explorer} = {\rm TRUE}$}{
$\vector{P} \leftarrow \vector{P} \cup \{\vector{u}\}$\;
}
$t \leftarrow t + 1$\;
}
%
\KwRet $\vector{P}$ for the decision making (Subsection \ref{sec:decision_making}) or benchmarking (Subsection \ref{sec:ada_postprocessing})\;
\caption{The ADA framework}
\label{alg:ada}
\end{algorithm}\DecMargin{0.5em}

\subsection{Normalization}
\label{sec:ada_normalization}

Since the objective functions of most real-world problems are differently scaled, the normalization method is mandatory.
In ADA, the normalization method depends on an EMOA to be combined.
For example, NSGA-III-ADA uses the intercept-based normalization method in NSGA-III.

If the original EMOA does not have a normalization method (e.g., MOEA/D-AGR), we use the following simple normalization method.
The objective vector $\vector{f}(\vector{x})$ is normalized using the approximated ideal point $\vector{z}^* = (z^*_1, ..., z^*_M)^{\rm T}$ and the worst point in the population $\vector{z}^{\rm worst} = (z^{\rm worst}_1, ..., z^{\rm worst}_M)^{\rm T}$.
The $i$-th element of the normalized objective vector $\vector{f}'(\vector{x})$ is given as: $f'_i(\vector{x}) = (f_i(\vector{x}) -z^*_i) /(z^{\rm worst}_i - z^*_i)$.




\subsection{Assignment operation}
\label{sec:ada_assignment}

The child $\vector{u}$ is assigned to the $j$-th subproblem based on its normalized objective vector (line 10 in Algorithm \ref{alg:ada}).
The assignment operation plays a crucial role in ADA to maintain the diversity of the population in the objective space.




Each EMOA has a different method to select the index $j$ of the subproblem to which $\vector{u}$ is assigned.
MOEA/D-AGR-ADA and eMOEA/D-ADA select the $j$-th subproblem with the minimum scalarizing function value $g(\vector{u}|\vector{w}_j)$ as in \eqref{eqn:moeadagr_selection}.
MOEA/D-DU-ADA, NSGA-III-ADA, and $\theta$-DEA-ADA assign $\vector{u}$ to the $j$-th subproblem with the minimum perpendicular distance between $\vector{f}'(\vector{u})$ and $\vector{w}_j$ as in \eqref{eqn:nsgaiii_selection}.
RVEA selects the subproblem with the minimum angle between $\vector{f}'(\vector{u})$ and the unit reference vector $\vector{v}_j$ as in \eqref{eqn:rvea_selection}.
ADA requires only a single index $j$ for $\vector{u}$ whereas MOEA/D-AGR, eMOEA/D, and MOEA/D-DU select $K$ subproblem indices from $\{1, ..., N\}$.




\subsection{Neighborhood criterion in the solution space}
\label{sec:ada_neighborhood}


ADA requires the neighborhood criterion in the solution space (line 11 in Algorithm \ref{alg:ada}).
While the objective space diversity is maintained by the assignment operation, the solution space diversity is controlled by the neighborhood criterion.


We use a simple relative distance-based neighborhood criterion presented in \cite{TanabeI18}.
First, the normalized Euclidean distance between each individual $\vector{x}$ in $\vector{P}$ and $\vector{u}$ is calculated in the solution space.
The upper and lower bounds for each decision variable of a problem are used for the normalization.
Then, all $\mu$ individuals in $\vector{P}$ are sorted based on their distance values in descending order.
If $\vector{x}$ is within the $L$ nearest individuals from $\vector{u}$, $\vector{x}$ is said to be a neighbor of $\vector{u}$ in the solution space.
$L$ is a control parameter in this neighborhood criterion.




In addition to the relative distance-based neighborhood criterion, any neighborhood criterion can be incorporated into ADA.
A number of niching methods have been proposed in the multi-modal single-objective optimization community \cite{LiEDE17}.
Modern niching methods include the adaptive radius-based method \cite{BirdL06} and the nearest-better clustering \cite{Preuss12}.
However, our preliminary results show that such sophisticated niching methods do not work well in ADA.
The main reason for the failure is that an appropriate parameter specification for such a modern niching method is difficult due to the difficulty in understanding the fitness landscape of an MOP \cite{KerschkeWPGDTE16}.
For this reason, we use the simple neighborhood criterion.

\subsection{Deletion operation}
\label{sec:ada_deletion}

\begin{figure}[t]
  \begin{center}
    \includegraphics[width=0.42\textwidth]{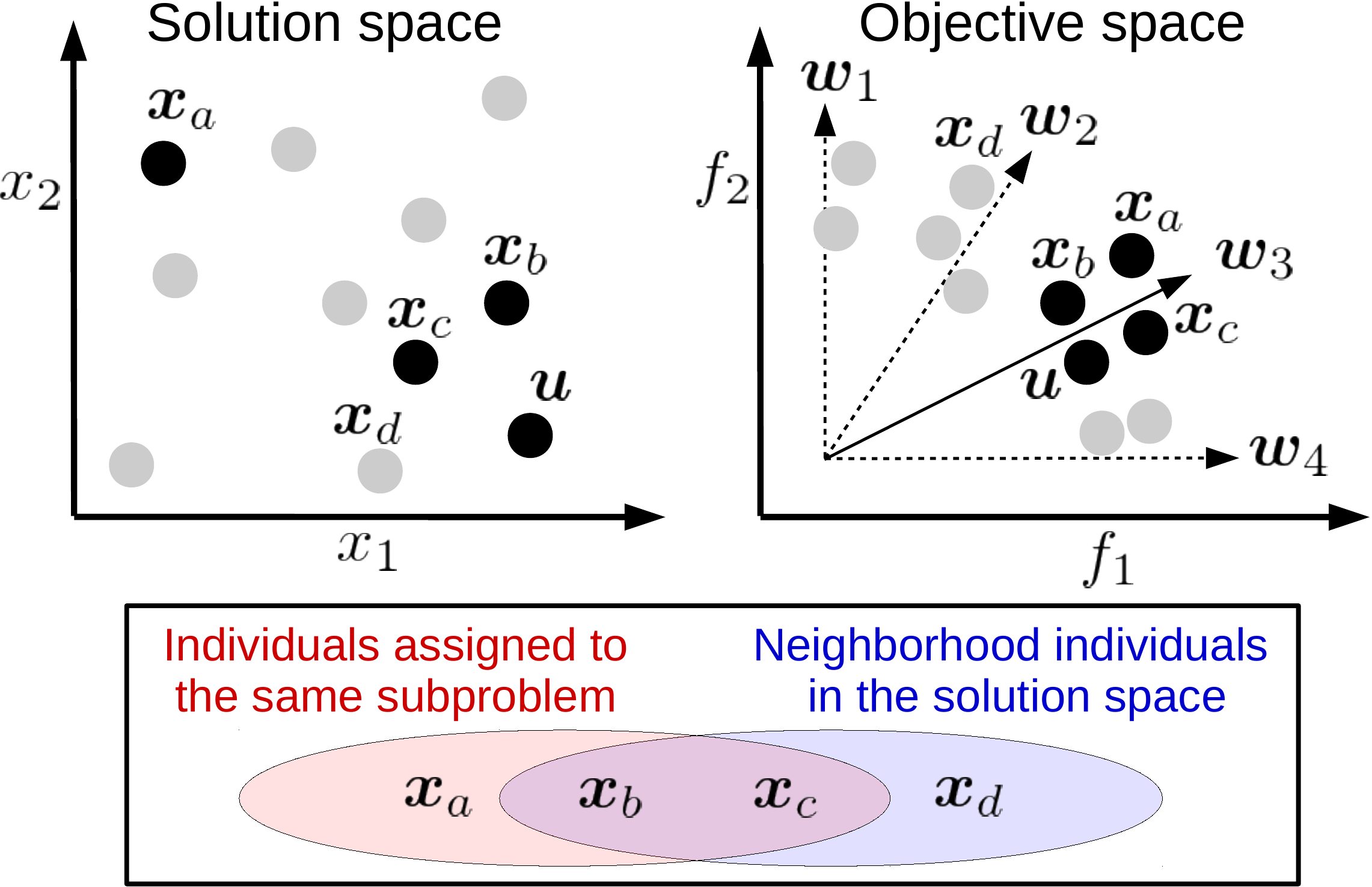}
    \caption{
\small
Example of $\vector{X}$. In this example, $\vector{X} = \{\vector{x}_b, \vector{x}_c\}$.
Since $\vector{x}_a$ and $\vector{x}_d$ are not compared to the child $\vector{u}$ in ADA, $\vector{x}_a$ and $\vector{x}_d$ survive to the next iteration independent of their quality.
}
\label{fig:example_X}
  \end{center}
\end{figure}

Let $\vector{X}$ be a set of individuals that are in the neighborhood of $\vector{u}$ in the solution space among the individuals assigned to the same $j$-th subproblem as $\vector{u}$ (line 11 in Algorithm \ref{alg:ada}).
Fig. \ref{fig:example_X} shows an example of $\vector{X}$.
In Fig. \ref{fig:example_X}, $\mu = 10$, $N=4$, $L=3$, and $\vector{u}$ has been assigned to the third-subproblem ($\vector{w}_3$).
A set of neighborhood individuals of $\vector{u}$ in the solution space is $\vector{Y} = \{\vector{x}_b, \vector{x}_c, \vector{x}_d\}$.
A set of individuals assigned to the third subproblem in the objective space is $\vector{Z} = \{\vector{x}_a, \vector{x}_b, \vector{x}_c\}$.
In this case, $\vector{X} = \vector{Y} \cap \vector{Z} = \{\vector{x}_b, \vector{x}_c\}$.
Whereas $\vector{x}_d$ is the neighborhood individual of $\vector{u}$, $\vector{x}_d$ has been assigned to the second subproblem ($\vector{w}_2$).
$\vector{x}_a$ has been assigned to the third subproblem with $\vector{u}$, but $\vector{x}_a$ is not in the neighborhood of $\vector{u}$.
Since $\vector{x}_a$ is dominated by $\vector{u}$ with respect to the two objectives in Fig. \ref{fig:example_X}, $\vector{x}_a$ is deleted in most EMOAs.
In contrast, $\vector{x}_a$ can survive in $\vector{P}$ in ADA.
This is because $\vector{x}_a$ is not a neighbor of $\vector{u}$ (and any other individual assigned to the third subproblem). 
Thus, $\vector{x}_a$ is not compared to $\vector{u}$. 
Although the quality of $\vector{x}_a$ is poor, $\vector{x}_a$ contributes to the solution space diversity of $\vector{P}$.




A paired comparison between $\vector{u}$ and each $\vector{x}$ in $\vector{X}$ is performed (lines 15--17 in Algorithm \ref{alg:ada}).
If $\vector{x}$ is evaluated as being worse than $\vector{u}$ (by the evaluation criterion explained in the next paragraph), $\vector{x}$ is deleted from $\vector{P}$.
The deletion of such $\vector{x}$ is reasonable since it is based on both the quality in the objective space and the diversity in the solution space.


The comparison criterion depends on the environmental selection in each original EMOA.
In MOEA/D-type algorithms, the comparison is based on the scalarizing function $g$.
MOEA/D-AGR-ADA, MOEA/D-DU-ADA, and eMOEA/D-ADA use $g^{\rm tch}$ in \eqref{eqn:tchebycheff-mul}, $g^{\rm dtch}$ in \eqref{eqn:tchebycheff-div}, and $g^{\rm msf}$ in \eqref{eqn:msf}, respectively.
If $g(\vector{x} | \vector{w}_j) \geq g(\vector{u} | \vector{w}_j)$ on the $j$-th subproblem, $\vector{x}$ is removed from $\vector{P}$.
In NSGA-III-ADA and $\theta$-DEA-ADA, the primary comparison between $\vector{x}$ and $\vector{u}$ is based on the Pareto-dominance relation.
In NSGA-III-ADA, ties are broken by the perpendicular distance between the normalized objective vector and $\vector{w}_j$.
In $\theta$-DEA-ADA, the tie-breaker is the $\theta$-dominance.
RVEA-ADA uses the APD value given in \eqref{eqn:apd}.

\subsection{Addition operation}
\label{sec:ada_addition}

The child $\vector{u}$ is added to the population $\vector{P}$ if either of the following two conditions is met (lines 18--19 in Algorithm \ref{alg:ada}).
One is that no individual exists in $\vector{X}$.
An empty $\vector{X}$ means that there is no neighborhood individual of $\vector{u}$ in $\vector{P}$ in the solution space (or the objective space).
If the first condition is met, $\vector{u}$ enters $\vector{P}$ without any comparison.
Although $\vector{u}$ with inferior quality is likely to enter $\vector{P}$, it helps $\vector{P}$ to maintain the solution space diversity (or the objective space diversity).
While a dominated individual is unlikely to survive to the next iteration in most EMOAs, it can remain in $\vector{P}$ due to the first criterion in the addition operation.

The other is that $\vector{u}$ performs better than at least one individual in $\vector{X}$ in the deletion operation (lines 16--17 in Algorithm \ref{alg:ada}).
Since $\vector{u}$ has good quality in its neighborhood in the solution space, it should be added to $\vector{P}$.

\begin{figure*}[t]
\newcommand{\widthvar}{0.235}
  \begin{center}
\subfloat[$\vector{P}$ in the obj. space]{\includegraphics[width=\widthvar\textwidth]{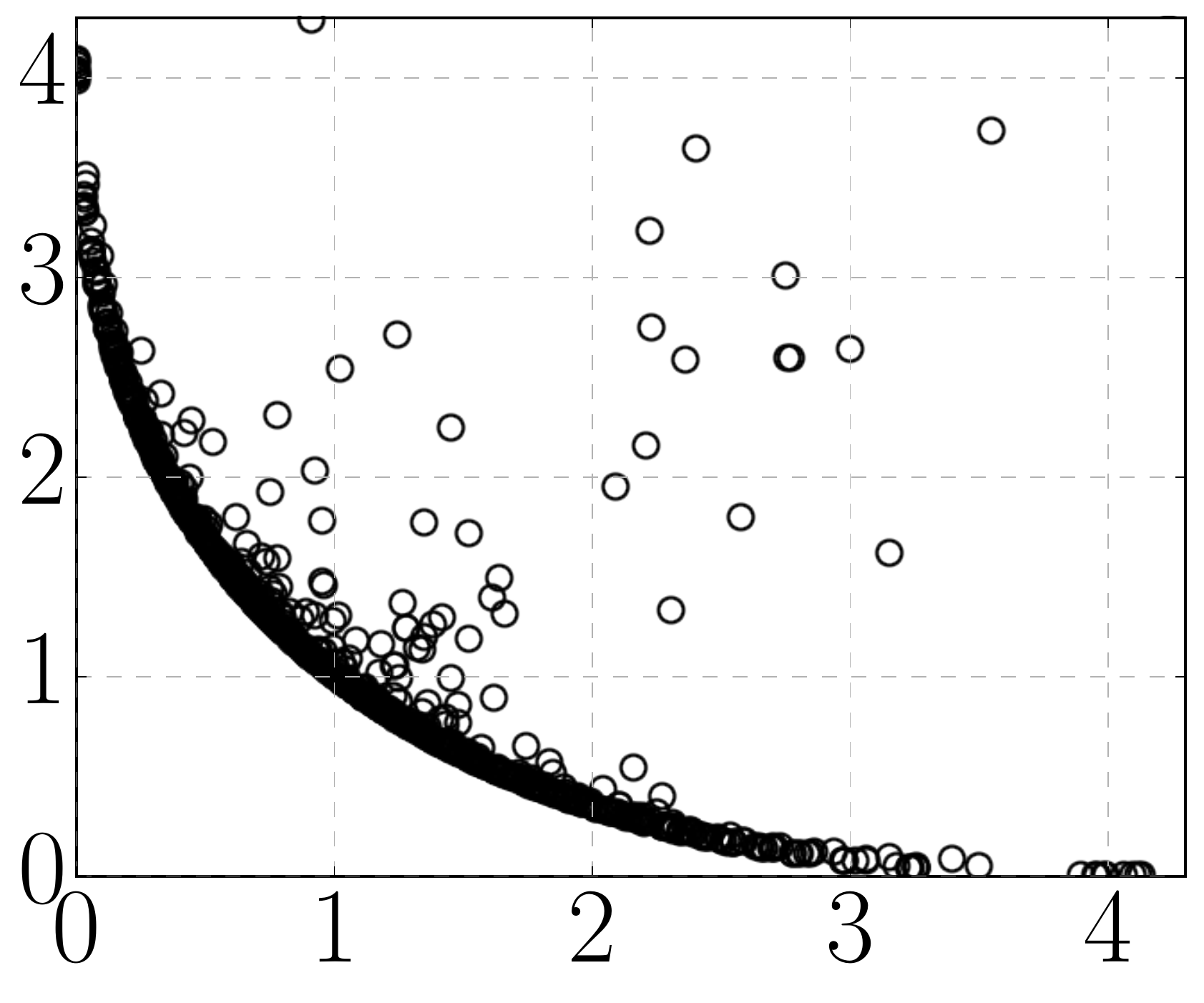}}
\subfloat[$\vector{A}^{\rm primary}$ in the obj. space]{\includegraphics[width=\widthvar\textwidth]{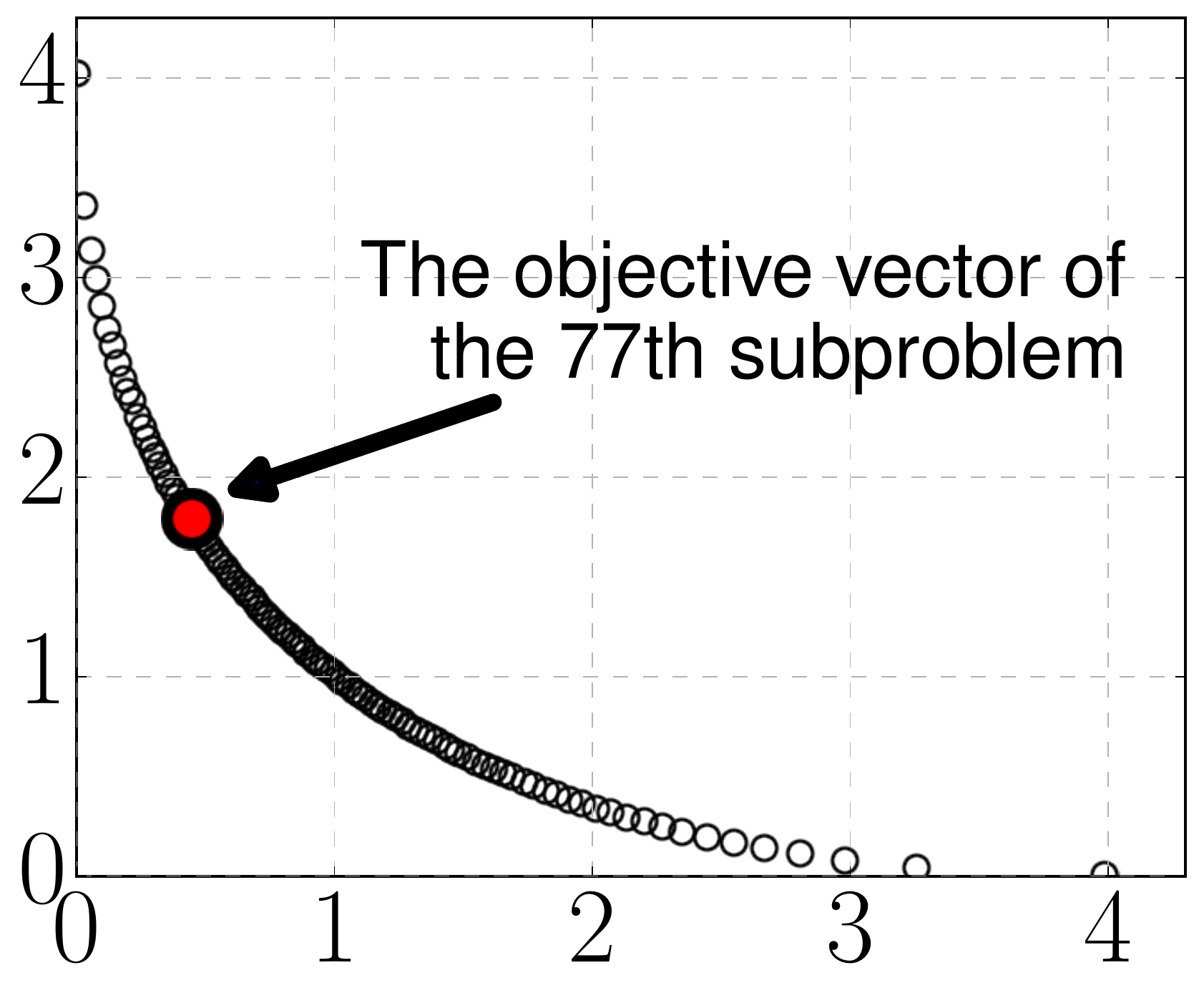}}
\subfloat[$\vector{A}^{\rm secondary}$ in the obj. space]{\includegraphics[width=\widthvar\textwidth]{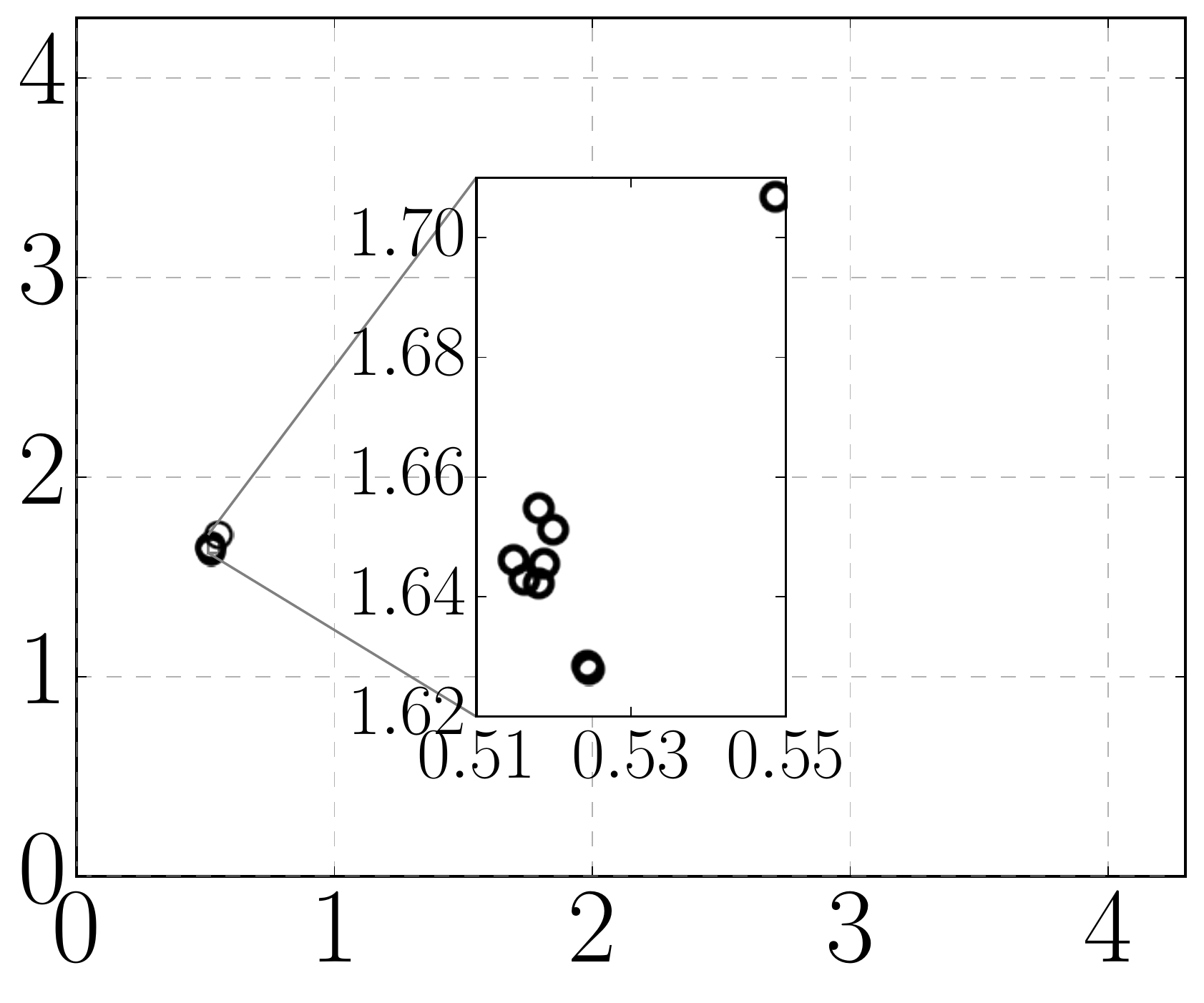}}
\subfloat[$\vector{A}^{\rm secondary}$ in the solution space]{\includegraphics[width=0.27\textwidth]{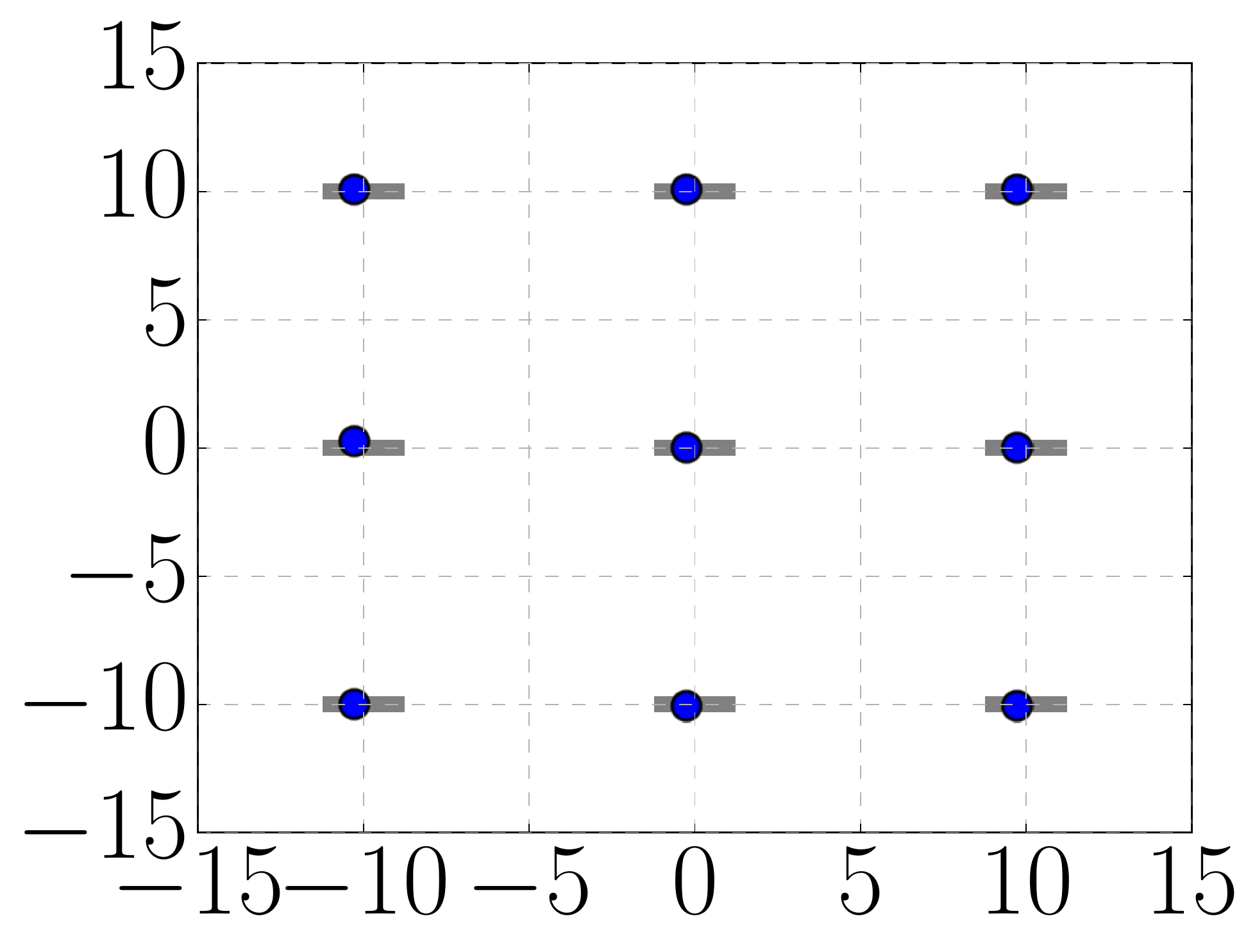}}
    \caption{
\small
Example of the decision making based on multiple equivalent solutions found by the proposed approach.
Solutions obtained by MOEA/D-AGR-ADA on SYM-PART1 are shown.
These figures show results of a single run with a median IGD$^+$ value among 31 runs.
In Fig. \ref{fig:example_decision_making} (a), (b), and (c), the x and y axes represent $f_1$ and $f_2$, respectively.
In Fig. \ref{fig:example_decision_making} (d), the x and y axes represent $x_1$ and $x_2$, respectively.
The nine gray lines in Fig. \ref{fig:example_decision_making} (d) are the equivalent Pareto optimal solution sets.
}
\label{fig:example_decision_making}
  \end{center}
\end{figure*}

\subsection{Decision making}
\label{sec:decision_making}

We introduce an effective decision making with a solution set found by an ADA-based algorithm.
Fig. \ref{fig:example_decision_making} shows an example of the decision making.
Fig. \ref{fig:example_decision_making} (a) exhibits the distribution of all objective vectors in the final population $\vector{P}$ of MOEA/D-AGR-ADA on the two-objective and two-variable SYM-PART1 problem \cite{RudolphNP07}.
As shown in Fig. \ref{fig:example_decision_making} (d), nine equivalent Pareto optimal solution sets are on the nine lines in SYM-PART1.
The experimental setting is described in Section \ref{sec:experimental_settings} later.
The decision making in ADA is the following two-phase method based on $\vector{A}^{\rm primary}$ and $\vector{A}^{\rm secondary}$.

\subsubsection{$\vector{A}^{\rm primary}$-based decision making}


Fig. \ref{fig:example_decision_making} (a) shows that $\vector{P}$ contains a large number of solutions.
This is undesirable for the decision maker, because she/he usually wants to examine a small number of well-distributed solutions in the objective space \cite{ZhangL07}.
To address this issue, the best solution for the $j$-th subproblem is selected from $\vector{P}$ based on the same criterion in the deletion operation ($j \in \{1, ..., N\}$).
Then, only non-dominated solutions are selected from the $N$ best solutions.
Let $\vector{A}^{\rm primary}$ be a set of the non-dominated solutions obtained by this procedure.
Fig. \ref{fig:example_decision_making} (b) shows non-dominated solutions in $\vector{A}^{\rm primary}$ in the objective space.
The decision maker examines objective vectors in $\vector{A}^{\rm primary}$, rather than $\vector{P}$.

When the decision maker wants to examine $N$ or less non-dominated solutions (e.g., only 10 solutions), other solution selection methods can be used to select $\vector{A}^{\rm primary}$ from $\vector{P}$.
If the number of objectives $M$ is less than $4$, efficient hypervolume-based selection approaches (e.g., \cite{BringmannFK14gecco}) are available.
If $M \geq 4$, computationally cheap distance-based selection methods (e.g., \cite{SinghBR18}) can be used.





\subsubsection{$\vector{A}^{\rm secondary}$-based decision making}

After the decision maker has examined $\vector{A}^{\rm primary}$ in the objective space, she/he selects the final solution $\vector{x}^{\rm final}$ from $\vector{A}^{\rm primary}$.
Suppose that the decision maker has selected a solution on the 77-th subproblem as $\vector{x}^{\rm final}$ in Fig. \ref{fig:example_decision_making} (b).
In ADA, she/he can examine other dissimilar solutions with similar quality to $\vector{x}^{\rm final}$.
Let $\vector{A}^{\rm secondary}$ be a set of all solutions assigned to the same 77-th subproblem with $\vector{x}^{\rm final}$. 
Figs. \ref{fig:example_decision_making} (c) and (d) show the distribution of all nine solutions in $\vector{A}^{\rm secondary}$ in the objective and solution spaces, respectively.
Although the nine solutions in $\vector{A}^{\rm secondary}$ have almost the same quality in the objective space, they are dissimilar in the solution space.
In addition to $\vector{x}^{\rm final}$, the decision maker can examine other candidates in $\vector{A}^{\rm secondary}$ based on her/his preference in the solution space.

\subsection{Two Post-processing methods for benchmarking}
\label{sec:ada_postprocessing}


Apart from the practical two-phase decision making method described in Subsection \ref{sec:decision_making}, we here consider benchmarking of ADA-based EMMAs.
The performance of EMOAs and EMMAs is evaluated using performance indicators.
However, a fair performance comparison is difficult between ADA-based EMMAs and other optimizers.
This is because the current population $\vector{P}$ in ADA can contain an unbounded number of solutions.
Most performance indicators cannot assess solution sets with different sizes in a fair manner \cite{IshibuchiSMN16a}.
IGD$^+$ \cite{IshibuchiMTN15} and IGDX \cite{ZhouZJ09} described in Section \ref{sec:experimental_settings} later are not exceptions.
Thus, ADA requires a method of selecting a constant number of solutions from $\vector{P}$ for benchmarking studies.



We introduce two post-processing methods that are used to evaluate the performance of ADA-based algorithms for MOPs and MMOPs, respectively.
In general, uniformly distributed solutions are unlikely to be uniformly distributed objective vectors due to a non-uniform mapping from the solution space to the objective space.
To address this issue, we use the two post-processing methods for the objective and solution spaces, respectively.
On the one hand, the same method of selecting $\vector{A}^{\rm primary}$ from $\vector{P}$ described in Subsection \ref{sec:decision_making} is used for performance indicators of MOPs (e.g., IGD \cite{CoelloS04}).
Since indicators of MOPs assess the distribution of a solution set in the objective space, the choice of $\vector{A}^{\rm primary}$ is reasonable.

On the other hand, the solution distance-based selection method presented in \cite{TanabeI18} is used for performance indicators for MMOPs (e.g., IGDX \cite{ZhouZJ09}).
Let us consider the task of selecting $N$ sparsely distributed solutions from all non-dominated solutions in $\vector{P}$.
Below, $D(\vector{x}, \vector{A})$ denotes the distance between a solution $\vector{x}$ and its nearest solution in a solution set $\vector{A}$ in the normalized solution space.
First, $\vector{A}^{\rm tertiary}$ is set to empty.
A solution is randomly selected from the non-dominated solution set and stored into $\vector{A}^{\rm tertiary}$.
Then, a solution with the maximum $D(\vector{x}, \vector{A}^{\rm tertiary})$ value is repeatedly added to $\vector{A}^{\rm tertiary}$ until $|\vector{A}^{\rm tertiary}| = N$.
Unlike $\vector{A}^{\rm primary}$ and $\vector{A}^{\rm secondary}$, $\vector{A}^{\rm tertiary}$ is used only for benchmarking of ADA-based algorithms.

\subsection{Applicability of ADA}
\label{sec:ada_applicability}



ADA is a framework to improve the performance of decomposition-based EMOAs for MMOPs.
We do not claim that ADA can be combined into any decomposition-based EMOAs.
Since ADA requires a method of assigning a child $\vector{u}$ to a subproblem, ADA is inapplicable to EMOAs with no assignment mechanism.
Such EMOAs include MOEA/D \cite{ZhangL07} and MOEA/D-DRA \cite{ZhangLL09}.
Also, the deletion operation performs the pairwise comparison independently from other individuals.
Thus, ADA is not applicable to EMOAs whose environmental selection is performed for all individuals in $\vector{P}$, such as MOEA/D-STM \cite{LiZKLW14} and VaEA \cite{XiangZLC16}.


In summary, ADA can be combined into EMOAs with an assignment mechanism and a pairwise comparison-based environmental selection.
The six EMOAs explained in Subsections \ref{sec:dmoea} and \ref{sec:rmoea} satisfy these conditions. 
In addition, ADA is applicable to I-DBEA \cite{AsafuddoulaRS15} and SPEA/R \cite{JiangY17a}.

Fortunately, even if a decomposition-based EMOA does not satisfy the above-mentioned two conditions, ADA can be applied to the EMOA after some modifications.
For example, since MOEA/D does not have the assignment method, we cannot directly combine ADA in MOEA/D.
However, MOEA/D can be easily modified by using any of the three assignment methods described in Subsection \ref{sec:ada_assignment}.
In fact, MOEA/D-AD proposed in \cite{TanabeI18} is an ADA-based MOEA/D with the assignment operation in \eqref{eqn:nsgaiii_selection}.
Thus, the applicability of ADA is not limited to only a few decomposition-based EMOAs.


\subsection{Originality of ADA}
\label{sec:ada_originality}

In addition to MOEA/D-AD \cite{TanabeI18}, a variant of MOEA/D for MMOPs is proposed in \cite{HuI18}.
The MOEA/D variant assigns $K$ individuals to each subproblem.
The fitness value is based on the PBI function value and two distance values in the solution space.
The main disadvantage of the MOEA/D variant in \cite{HuI18} is the difficulty in finding a proper $K$ value.
Since the number of equivalent Pareto optimal solution subsets is unknown a priori, fine-tuning of $K$ is necessary for a given problem.
Although a multi-start decomposition-based approach is proposed in \cite{RudolphNP07}, it has a similar disadvantage.
In contrast, ADA adaptively adjusts the number of individuals assigned to each subproblem.
Thus, ADA does not require a problem-dependent parameter such as $K$.


TriMOEA-TA\&R \cite{LiuYG18} consists of various advanced components, including the convergence and diversity archives-based strategy as in Two\_Arch2 \cite{WangJY15}, the decision variable-decomposition method in MOEA/DVA \cite{MaLQWLJYG16}, and the angle-based individual assignment in RVEA \cite{ChenJOS16}.
ADA and TriMOEA-TA\&R are similar in that they handle diversity in the objective and solution spaces in a two-phase manner.
While TriMOEA-TA\&R uses an {\em absolute} distance-based neighborhood criterion with $\sigma_{\rm niche}$, ADA uses the {\em relative} distance-based neighborhood criterion with $L$.
ADA uses a much simpler mating scheme, and it does not use any decomposition method of decision variables.
ADA also uses the adaptive population sizing strategy to handle equivalent solutions, as mentioned above.
Whereas TriMOEA-TA\&R is an algorithm for MMOPs with distance-related variables, ADA is a general framework for various MMOPs.

\section{Experimental settings}
\label{sec:experimental_settings}

\begin{table}[t]
\begin{center}
  \caption{\small Properties of multi-modal multi-objective test problems, where $M$, $D$, and $O$ denote the number of objectives, design variables, and equivalent Pareto optimal solution subsets, respectively.}
{\footnotesize
  \label{suptab:mmop_test_problems}
\scalebox{1}[1]{ 
\begin{tabular}{lccccc}
\midrule
Test problems & $M$ & $D$ & $O$\\ 
\toprule
Two-On-One \cite{PreussNR06} and SSUF1,3\cite{LiangYQ16} & 2  & 2 & 2 \\\midrule
SYM-PART1--3 \cite{RudolphNP07} & 2  & 2 & 9\\\midrule
Polygon  \cite{IshibuchiHTN10} & Any  & 2 & Any \\\midrule
Omni-test \cite{DebT08} & 2 & Any & $3^D$ \\\midrule
\end{tabular}
}
}
\end{center}
\end{table}

\subsection{Test problems}
\label{sec:test_problems}

We use the following eight multi-modal multi-objective test problems: the Two-On-One problem \cite{PreussNR06}, the three SYM-PART problems \cite{RudolphNP07}, the two SSUF problems \cite{LiangYQ16}, the Polygon problem \cite{IshibuchiHTN10}, and the Omni-test problem \cite{DebT08}.
Table \ref{suptab:mmop_test_problems} shows their properties, including the number of objectives $M$, the number of decision variables $D$, and the number of equivalent Pareto optimal solution sets $O$.



Two-On-One has two equivalent Pareto optimal solution sets that are symmetrical with respect to the origin.
Equivalent Pareto optimal solution sets are on the nine lines in SYM-PART1, as shown in Fig. \ref{fig:example_decision_making}.
The nine lines are rotated in SYM-PART2.
In addition, the nine lines are distorted in SYM-PART3.
SSUF1 and SSUF3 have two symmetrical Pareto optimal solution sets.
%
We evaluate the scalability of EMMAs to $M$ using Polygon.
We set $M$ of Polygon as follows: $M \in \{3, 5, 8, 10\}$.
Although $O$ can be any number in Polygon, it was set to be nine.
We investigate the scalability of EMMAs to $D$ and $O$ using Omni-test.
We set $D$ as follows: $D \in \{2, 3, 5, 8, 10\}$.
$O$ increases exponentially with increased $D$ in Omni-test.
In summary, we use 15 test problem instances.



HPS \cite{ZhangSA17} and MMMOP \cite{LiuYG18} have been recently proposed.
However, HPS and MMMOP have so-called ``distance-related'' variables that affect only the distance between the objective vector and the Pareto front.
For this reason, we do not mainly use HPS and MMMOP for our benchmarking study.
We use HPS and MMMOP only in Subsections \ref{sec:hps_mmmop} and \ref{sec:comparison_three_methods_uea}.

\subsection{Performance indicators} 
\label{sec:performance_indicators}


Below, $\vector{A}$ is a set of solutions obtained by an EMMA.
$\vector{A}^*$ is also a set of reference solutions in the Pareto optimal solution set.
The size of $\vector{A}^*$ was set to $5\,000$.
For $\vector{A}^*$ of each problem, $5\,000$ solutions were selected from randomly generated $10\,000$ Pareto-optimal solutions using the distance-based solution selection method \cite{TanabeI18} (Subsection \ref{sec:ada_postprocessing}).


We use IGD$^+$ \cite{IshibuchiMTN15} to evaluate $\vector{A}$ in terms of both convergence to the Pareto front and diversity in the objective space:
{\small
\begin{align}
\label{eqn:igd}
      {\rm IGD^+} (\vector{A}) &= \frac{1}{|\vector{A}^*|} \left(\sum_{\vector{z} \in \vector{A}^*} \min_{\vector{x} \in \vector{A}} \Bigl\{ d\bigl(\vector{f}(\vector{x}), \vector{f}(\vector{z})\bigr) \Bigr\} \right),
\end{align}
}%
where $d(\vector{a}, \vector{b}) = \sqrt{\sum^M_{i=1} \bigl(\max\{a_i - b_i, 0\}\bigr)^2}$.
IGD$^+$ is a modified version of IGD \cite{CoelloS04}.
While the original IGD is Pareto non-compliant, IGD$^+$ is weakly Pareto compliant.



We evaluate how well $\vector{A}$ approximates the Pareto-optimal solution set in the solution space using IGDX \cite{ZhouZJ09}:
{\small
\begin{align}
\label{eqn:igdxx}
{\rm IGDX} (\vector{A}) &= \frac{1}{|\vector{A}^*|} \left(\sum_{\vector{z} \in \vector{A}^*} \min_{\vector{x} \in \vector{A}} \Bigl\{ {\rm ED} \bigl(\vector{x}, \vector{z} \bigr) \Bigr\} \right),
 \end{align}
}%
where ${\rm ED}(\vector{a}, \vector{b})$ is the Euclidean distance between $\vector{a}$ and $\vector{b}$.




EMOAs that can find $\vector{A}$ with small IGD$^+$ and IGDX values are efficient multi-objective optimizers and multi-modal multi-objective optimizers, respectively.
In the ADA-based algorithms, $\vector{A}^{\rm primary}$ and $\vector{A}^{\rm tertiary}$ explained in Subsection \ref{sec:ada_postprocessing} are used for the IGD$^+$ and IGDX calculations, respectively.

\subsection{Average performance score}
\label{sec:settings_aps}

We use the average performance score (APS) \cite{BaderZ11} in order to aggregate results on various problems.
Suppose that $n$ algorithms $A_1, ..., A_n$ are compared for a problem instance based on the indicator values obtained in multiple runs.
For each $i \in \{1, ..., n\}$ and $ j \in \{1, ..., n\} $ $\setminus \{i\}$, if $A_j$ significantly outperforms $A_i$ using the Wilcoxon rank-sum test with $p < 0.001$, then $\delta_{i,j} = 1$; otherwise, $\delta_{i,j} = 0$.
The score $P(A_i)$ is defined as follows: $P(A_i) = \sum^{n}_{ j \in \{1, ..., n\} \backslash \{i\}} \delta_{i,j}$.
The score $P(A_i)$ represents the number of algorithms that outperform $A_i$. 
The APS value of $A_i$ is the average of the $P(A_i)$ values for all problem instances.
A small APS value of $A_i$ indicates that $A_i$ performs well among $n$ algorithms.


\subsection{EMMAs and EMOAs}
\label{sec:settings_EMMAs}


We examine the performance of the six ADA-based EMMAs: MOEA/D-AGR-ADA, MOEA/D-DU-ADA, eMOEA/D-ADA, NSGA-III-ADA, $\theta$-DEA-ADA, and RVEA-ADA.
We implemented the ADA-based algorithms using jMetal \cite{DurilloN11}.
Their source codes can be downloaded from the supplementary website ({\url{https://sites.google.com/view/mmoada}).
We compare the ADA-based EMMAs to their original EMOAs: MOEA/D-AGR \cite{WangZZGJ16}, MOEA/D-DU \cite{YuanXWZY16}, eMOEA/D \cite{JiangYWL18}, NSGA-III \cite{DebJ14}, $\theta$-DEA \cite{YuanXWY16}, and RVEA \cite{ChenJOS16}.
Our implementations of the six original EMOAs were based on their corresponding articles, except for MOEA/D-AGR.
We replaced the DE operator with SBX in MOEA/D-AGR to remove the effect of variation operators.
For details, refer to the corresponding articles.




  
The number of maximum function evaluations was $30\,000$.
$31$ runs were performed for each test problem.
A set of weight/reference vectors were generated using the simplex-lattice design method for $M < 6$ and its two-layered version \cite{DebJ14} for $M \geq 6$.
The number of the weight/reference vectors $N$ was $100$, $105$, $210$, $156$, and $230$, for $M=2$, $3$, $5$, $8$, and $10$, respectively.
The SBX crossover and the polynomial mutation were used in all methods, including MOEA/D-AGR.
Their control parameters were set as follows: $p_c = 1$, $\eta_c = 20$, $p_m = 1/D$, and $\eta_m = 20$.
According to the analysis presented in \cite{TanabeI18}, $L$ of the neighborhood criterion in ADA was set to $L=\lfloor 0.1 \mu \rfloor$.
For example, $L=201$ when $\mu = 2\,019$.
Other parameters were set according to the corresponding references.

\definecolor{c1}{RGB}{192,192,192}

\section{Experimental results}
\label{sec:experimental_results}

This section shows performance analysis of the six ADA-based EMMAs.
Subsection \ref{sec:comparison_ada} describes the effect of ADA on the six EMOAs.
Subsection \ref{sec:comparison_three_methods} compares the six ADA-based EMMAs to state-of-the-art EMMAs.
Subsection \ref{sec:results_popsize} discusses the adaptive population sizing in ADA.
Subsection \ref{sec:impact_l} investigates the influence of the $L$ value on the performance of the six ADA-based EMMAs.
Subsection \ref{sec:hps_mmmop} examines the performance of the six ADA-based EMMAs on test problems with distance-related variables.
Subsection \ref{sec:comparison_three_methods_uea} presents a comparison with the state-of-the-art EMMAs using an unbounded external archive.
Subsection \ref{sec:runtime} discusses the runtime of the ADA-based algorithms.



\begin{table}[t!]
\renewcommand{\arraystretch}{1}
\centering
\caption{\small Results of the six EMOAs and their ADA versions on the 15 test problem instances.
Tables (a) and (b) show the APS values of the algorithms for IGD$^+$ and IGDX, respectively.
AGR and DU stand for MOEA/D-AGR and MOEA/D-DU, respectively.
}
  \label{tab:comparison}
  \subfloat[IGD$^+$]{
    {\scriptsize
      \scalebox{0.9}[1]{
        \begin{tabular}{|l|c|c|}                  
          \hline
          & Orig. & ADA\\\hline
AGR-ADA & \cellcolor{c1}0.0 (1)  & 1.0 (2) \\ \hline
DU-ADA  & \cellcolor{c1}0.0 (1)  & 0.8 (2) \\ \hline
eMOEA/D-ADA & \cellcolor{c1}0.2 (1)  & 0.7 (2) \\ \hline
NSGA-III-ADA & \cellcolor{c1}0.0 (1)  & 0.8 (2) \\ \hline
$\theta$-DEA-ADA & \cellcolor{c1}0.1 (1)  & 0.6 (2) \\ \hline
RVEA-ADA & 0.5 (2)  & \cellcolor{c1}0.3 (1) \\ \hline
        \end{tabular}
      }
    }
  }
  \subfloat[IGDX]{
    {\scriptsize
      \scalebox{0.97}[1]{
        \begin{tabular}{|l|c|c|}                  
          \hline
          & Orig. & ADA\\\hline
AGR-ADA & 1.0 (2)  & \cellcolor{c1}0.0 (1) \\ \hline
DU-ADA & 0.9 (2)  & \cellcolor{c1}0.0 (1) \\ \hline
eMOEA/D-ADA & 1.0 (2)  & \cellcolor{c1}0.0 (1) \\ \hline
NSGA-III-ADA & 1.0 (2)  & \cellcolor{c1}0.0 (1) \\ \hline
$\theta$-DEA-ADA & 1.0 (2)  & \cellcolor{c1}0.0 (1) \\ \hline
RVEA-ADA & 1.0 (2)  & \cellcolor{c1}0.0 (1) \\ \hline
        \end{tabular}
      }
    }
  }
\end{table}

\subsection{Effect of ADA}
\label{sec:comparison_ada}


Tables \ref{tab:comparison} (a) and (b) show the paired comparison of each EMOA and its ADA version on the 15 test problem instances in terms of IGD$^+$ and IGDX, respectively.
Table \ref{tab:comparison} shows only the APS value of each algorithm at the final iteration.
Table S.1 in the supplementary file shows detailed results.
As described in Subsection \ref{sec:ada_postprocessing}, only $N$ solutions in $\vector{A}^{\rm primary}$ and $\vector{A}^{\rm tertiary}$ were used for the IGD$^+$ and IGDX calculations.
Thus, all algorithms are compared under the same number of solutions.
We set the $p$ value to $0.001$ for the Wilcoxon rank-sum test.
$M$-Polygon is the $M$-objective Polygon problem, and $D$-Omni-test is the $D$-variable Omni-test problem.
Below, we describe the results of IGD$^+$ and IGDX. 



\subsubsection{IGD$^+$}
Table \ref{tab:comparison} (a) shows that the original EMOAs outperform their ADA-based EMMAs regarding IGD$^+$ (except for RVEA-ADA).
In general, EMMAs perform slightly worse than EMOAs for multi-objective optimization \cite{ShirPNE09,YueQL17}.
While EMOAs aim to find a good approximation of the Pareto front, EMMAs attempt to locate all Pareto optimal solutions.
For this reason, the performance of the ADA versions regarding IGD$^+$ is worse than that of their original EMOAs.

However, as shown in Table S.1 in the supplementary file, the IGD$^+$ value of the ADA versions is only $3.57$ times worse than that of their original EMOAs even in the worst case.
Also, the ADA versions perform better than their original EMOAs on some problems.
For example, eMOEA/D-ADA obtains better IGD$^+$ values than eMOEA/D on SYM-PART2, SSUF3, and 10-Polygon.
As discussed in \cite{UlrichBZ10}, a mechanism for maintaining the solution space diversity can help the ADA versions to find high-quality solutions.

%


\subsubsection{IGDX}
%

Table \ref{tab:comparison} (b) shows that the ADA-based EMMAs significantly outperform their original EMOAs.
As shown in Table S.1, the IGDX value of the ADA versions is $141.58$ times better than that of their original EMOAs in the best case.
These results indicate that ADA improves the performance of the six decomposition-based EMOAs for MMOPs.

Table S.1 shows that the ADA-based EMMAs work well on Polygon with $M \in \{3, 5, 8, 10\}$ and Omni-test with $D \in \{2, 3, 5, 8, 10\}$.
Since the number of equivalent Pareto optimal solution sets $O$ increases exponentially with increased $D$ in Omni-test, the results show that ADA can also handle problems with a large $O$ value.
In summary, ADA has a sufficient scalability to $M$, $D$, and $O$.




\begin{figure}[t]
  \newcommand{\widthvar}{0.162}
  \begin{center} 
    %
\subfloat[eMOEA/D]{\includegraphics[width=\widthvar\textwidth]{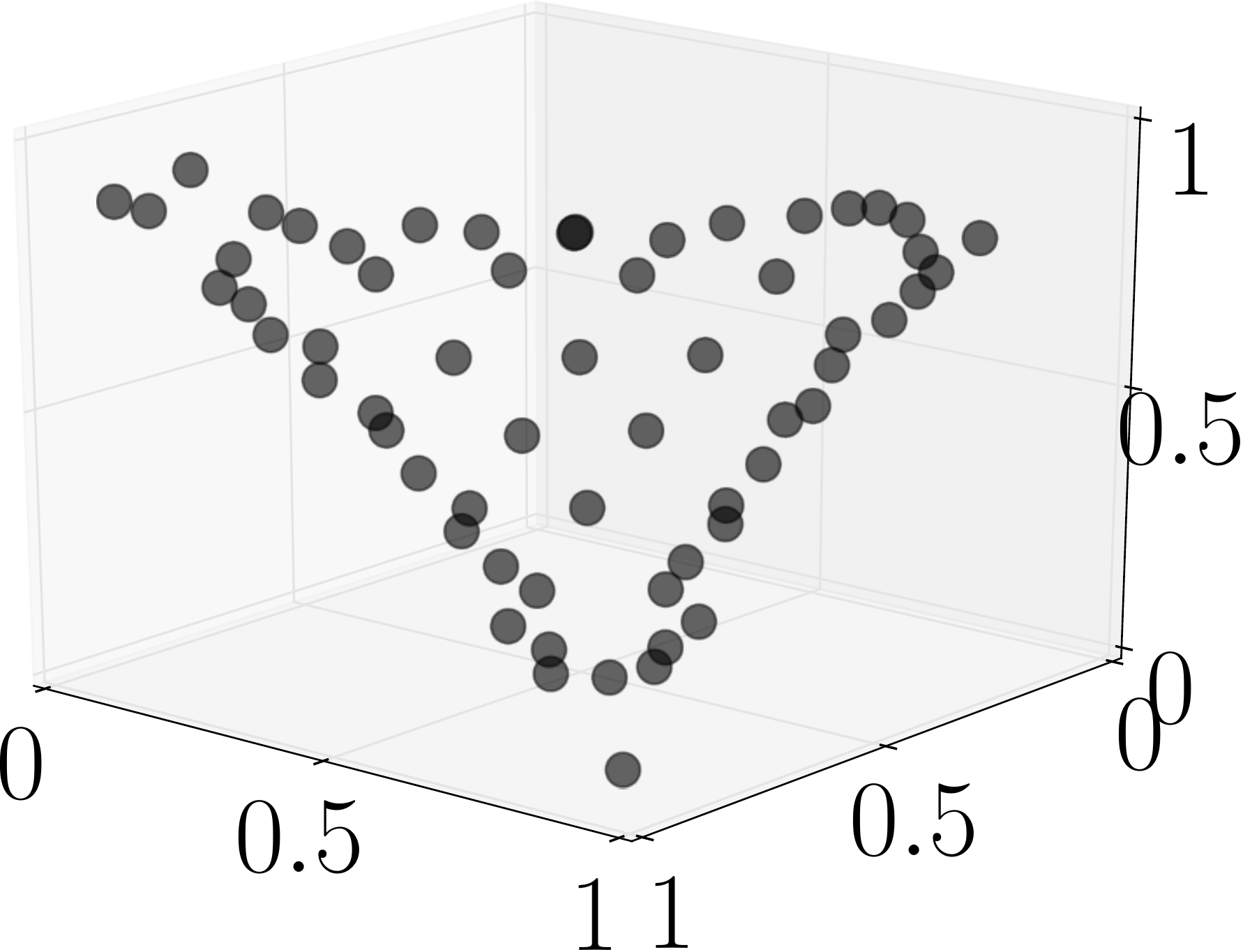}}
\subfloat[NSGA-III]{\includegraphics[width=\widthvar\textwidth]{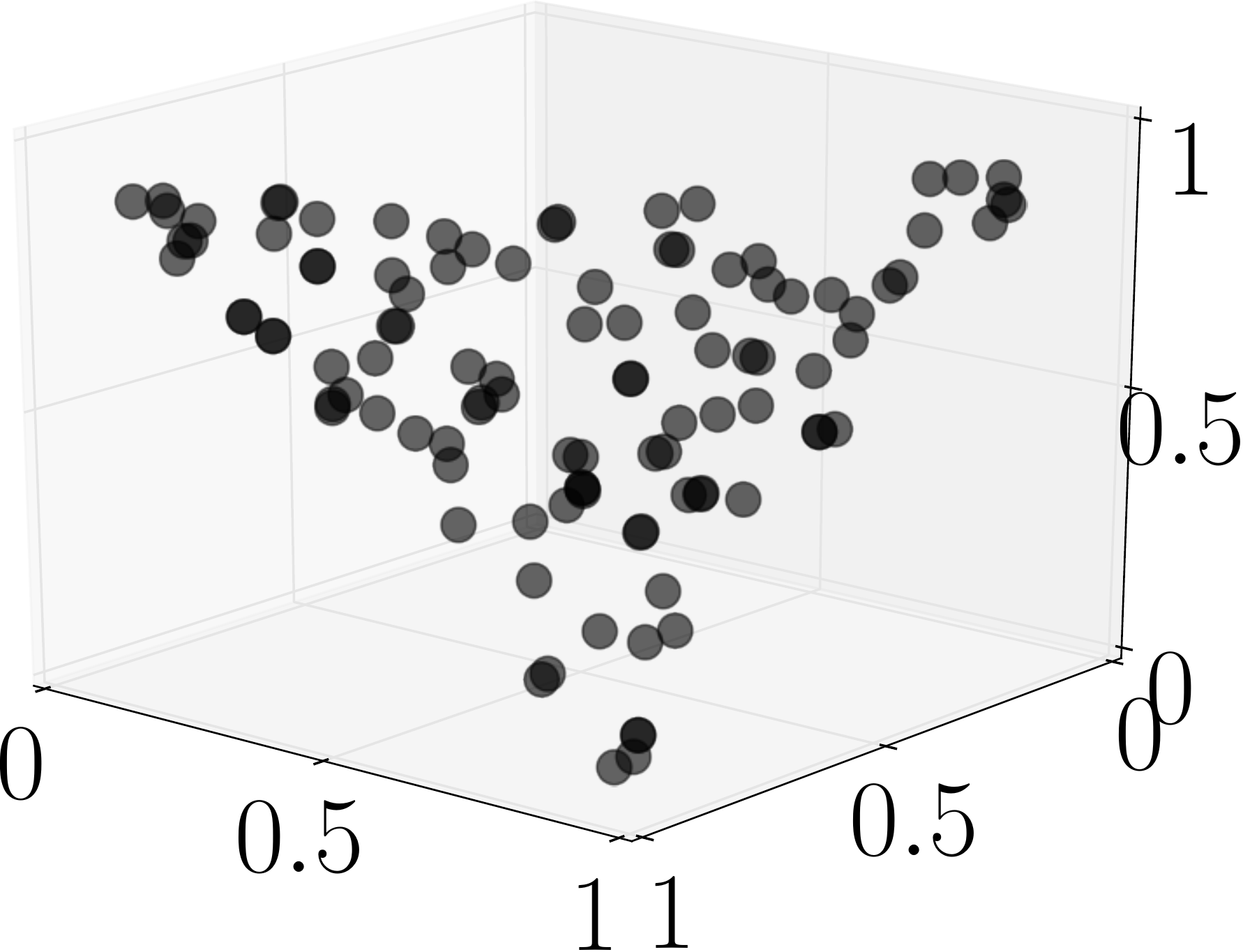}}
\subfloat[RVEA]{\includegraphics[width=\widthvar\textwidth]{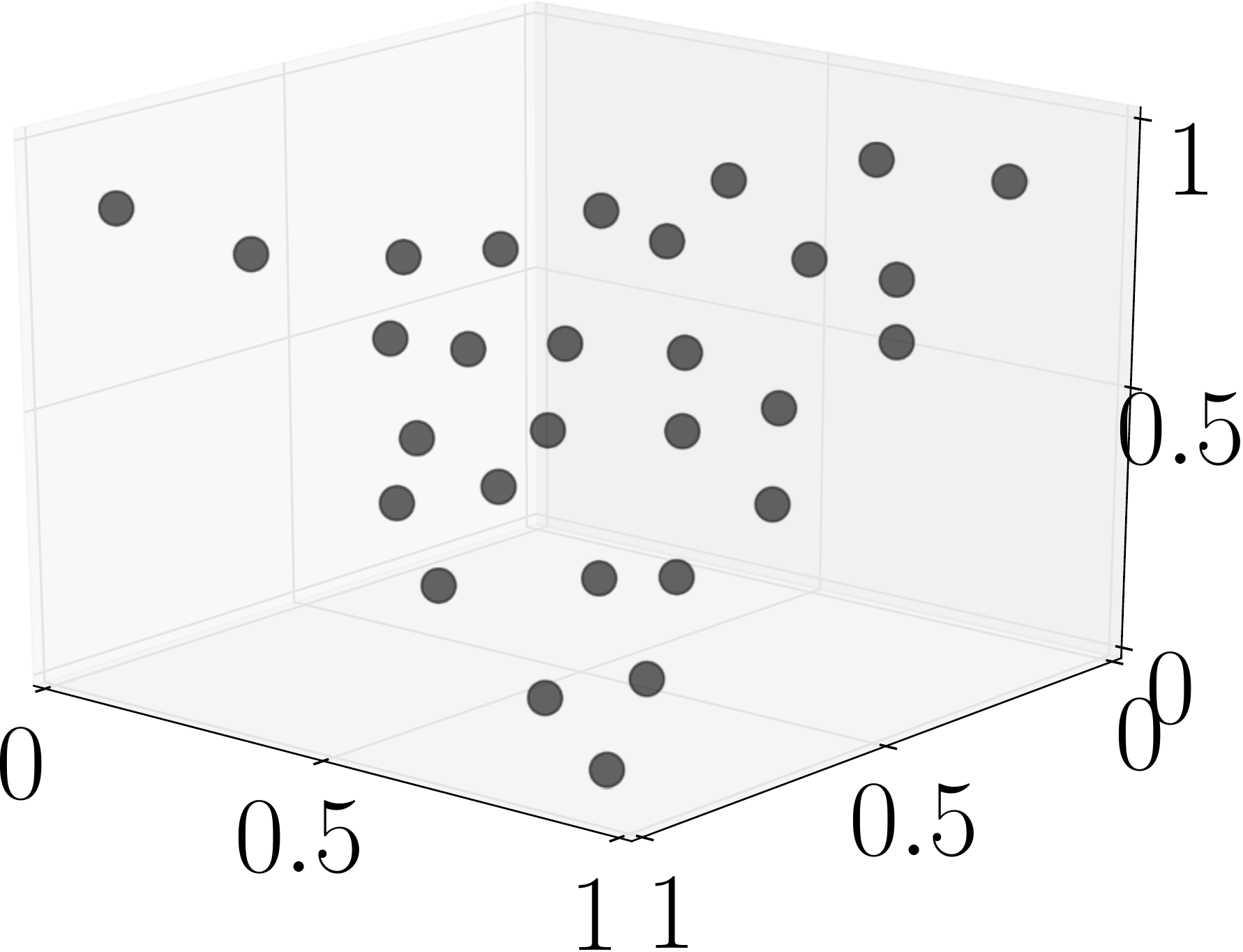}}
\\
\subfloat[eMOEA/D-ADA]{\includegraphics[width=\widthvar\textwidth]{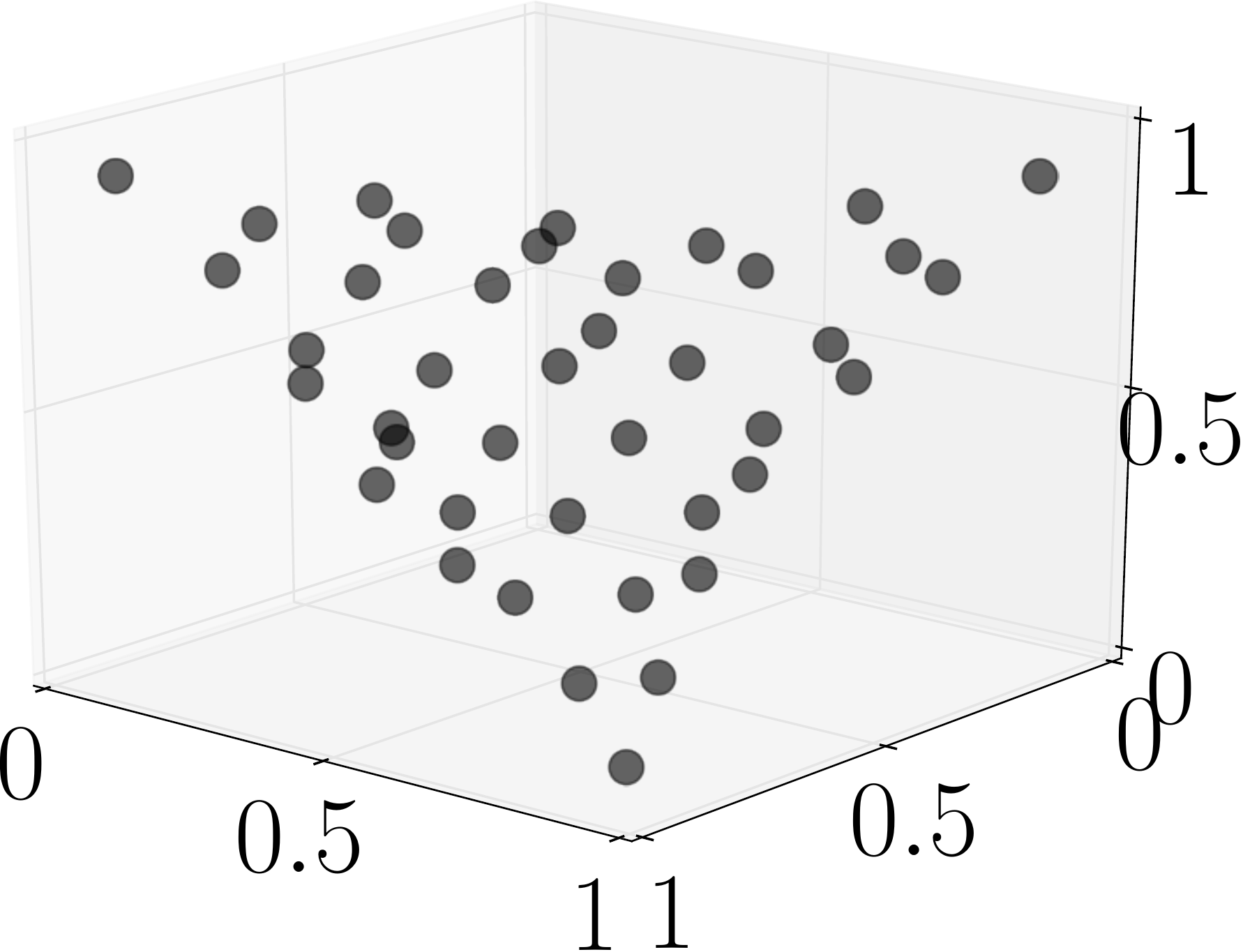}}
\subfloat[NSGA-III-ADA]{\includegraphics[width=\widthvar\textwidth]{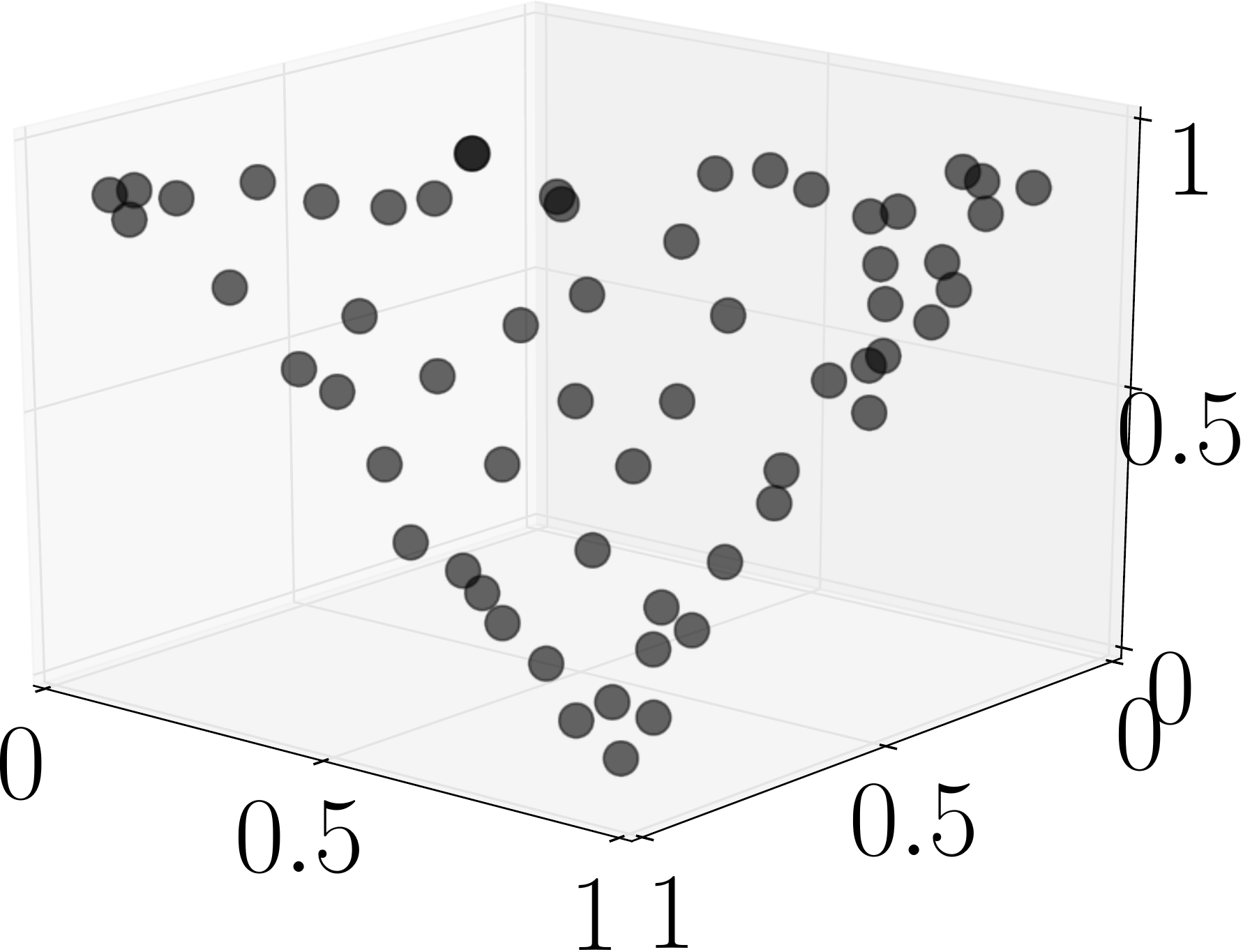}}
\subfloat[RVEA-ADA]{\includegraphics[width=\widthvar\textwidth]{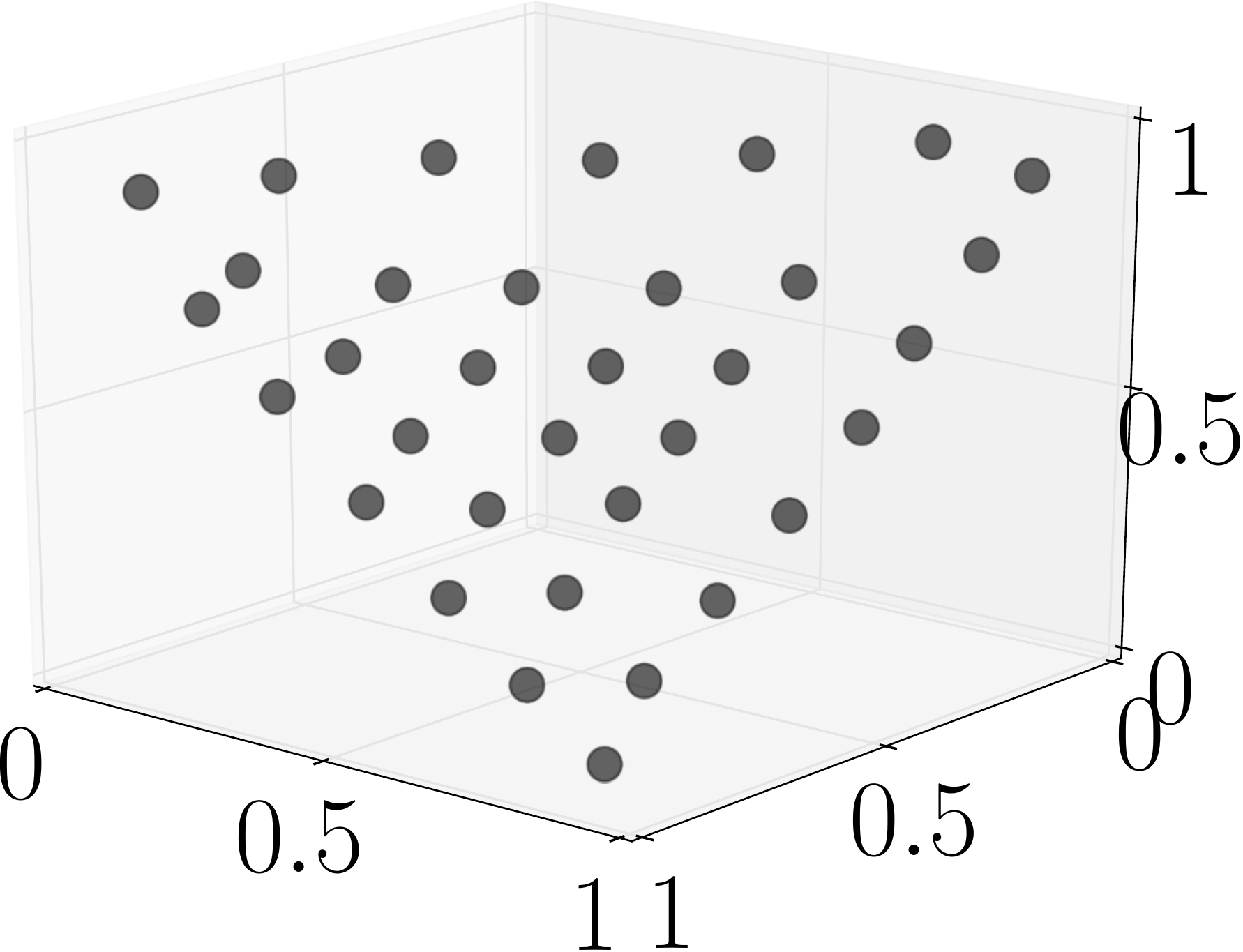}}
\caption{
\small
Distribution of non-dominated solutions found by each method in the objective space on the three-objective Polygon problem.
The x, y, and z axes represent $f_1$, $f_2$, and $f_3$, respectively.
}
\label{fig:3polygon_obj}
  \end{center}
%
  \begin{center} 
    %
\subfloat[eMOEA/D]{\includegraphics[width=\widthvar\textwidth]{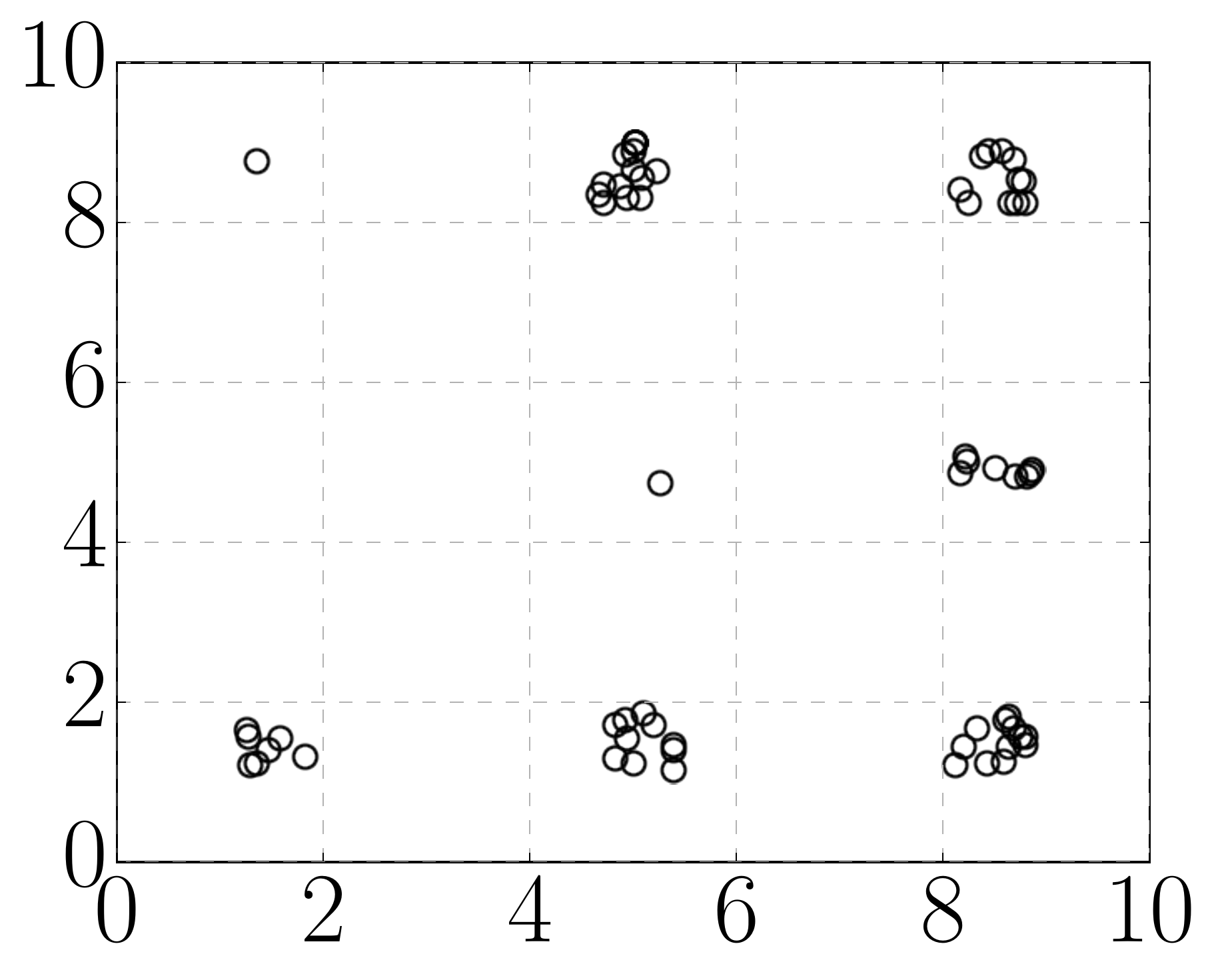}}
\subfloat[NSGA-III]{\includegraphics[width=\widthvar\textwidth]{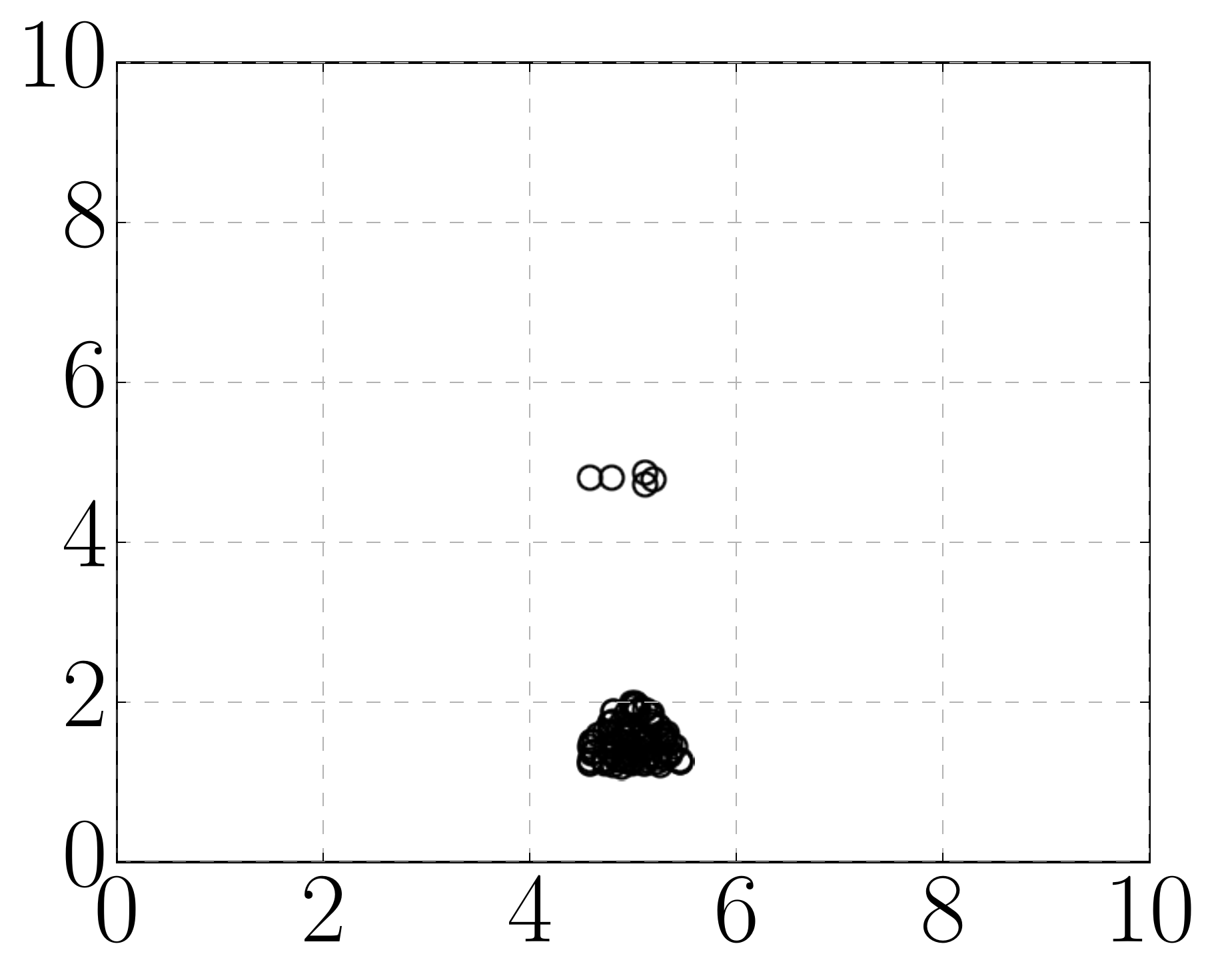}}
\subfloat[RVEA]{\includegraphics[width=\widthvar\textwidth]{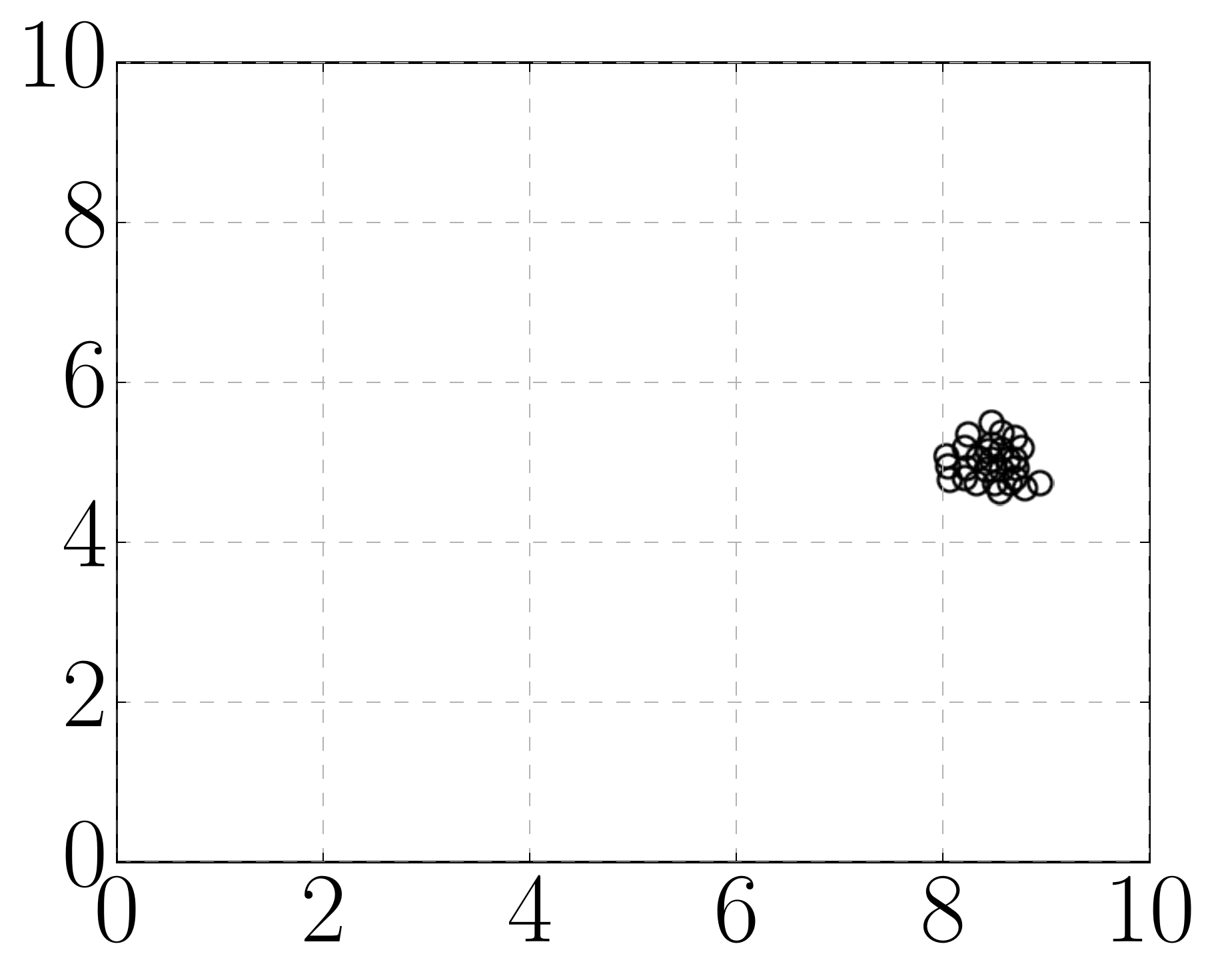}}
\\
\subfloat[eMOEA/D-ADA]{\includegraphics[width=\widthvar\textwidth]{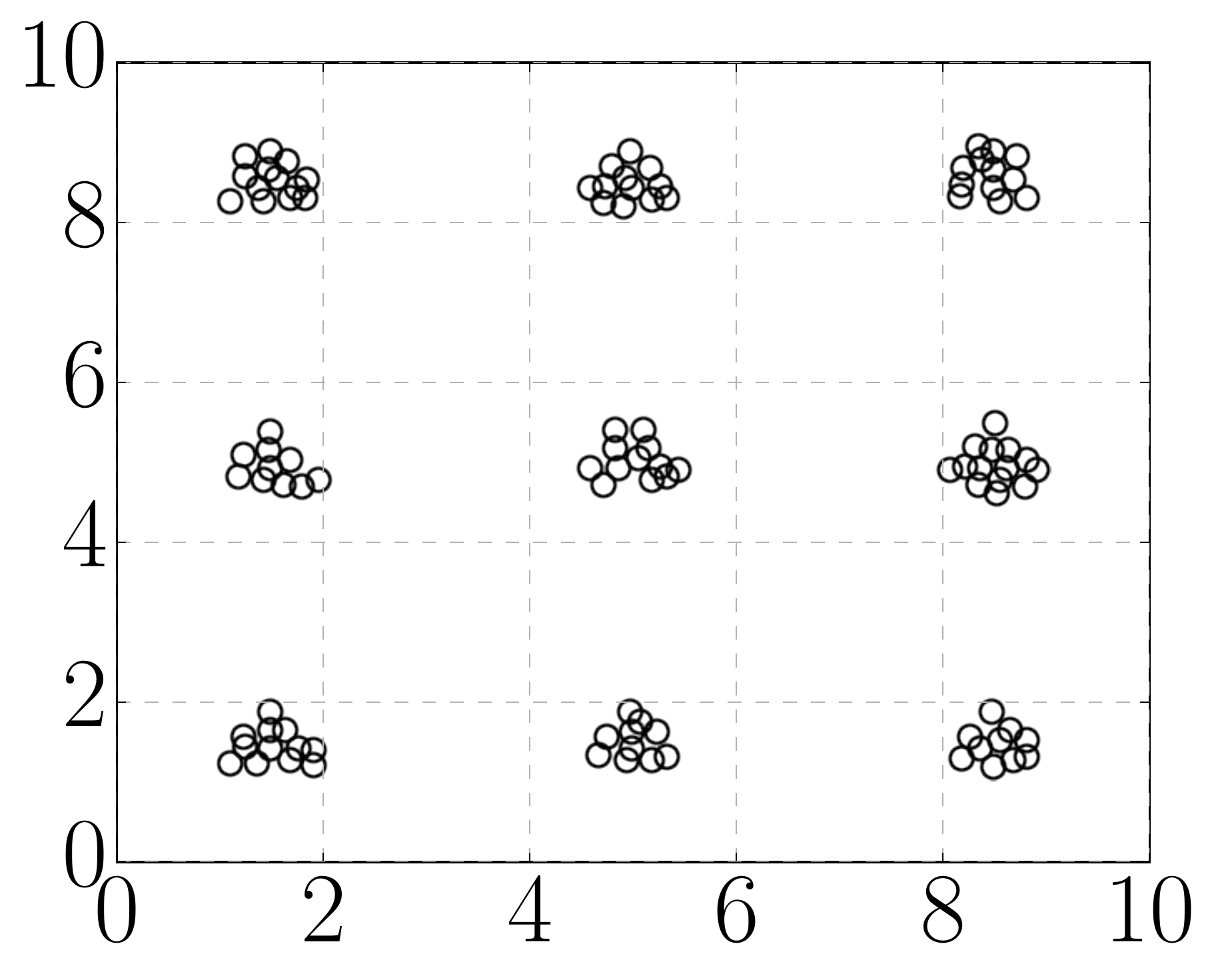}}
\subfloat[NSGA-III-ADA]{\includegraphics[width=\widthvar\textwidth]{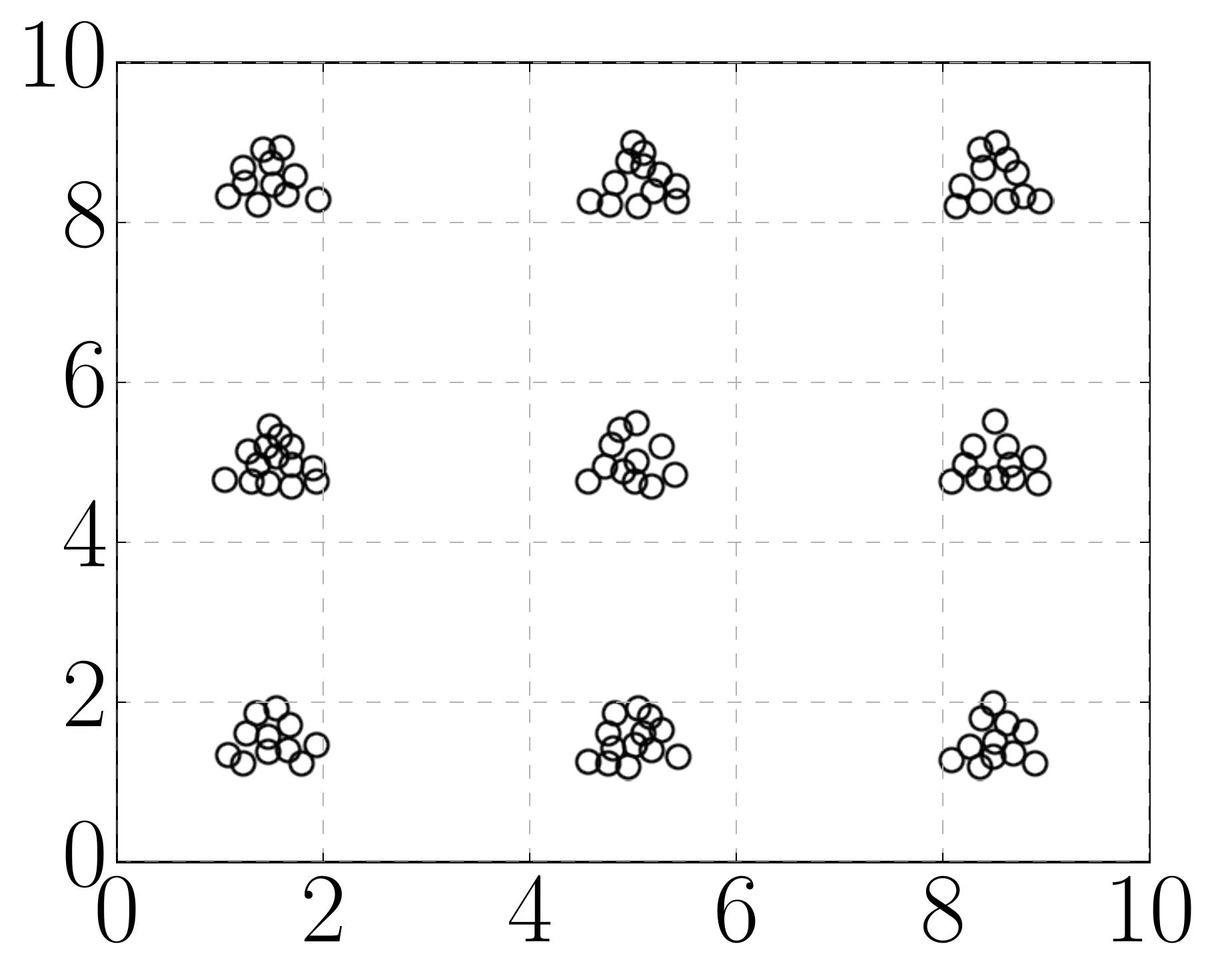}}
\subfloat[RVEA-ADA]{\includegraphics[width=\widthvar\textwidth]{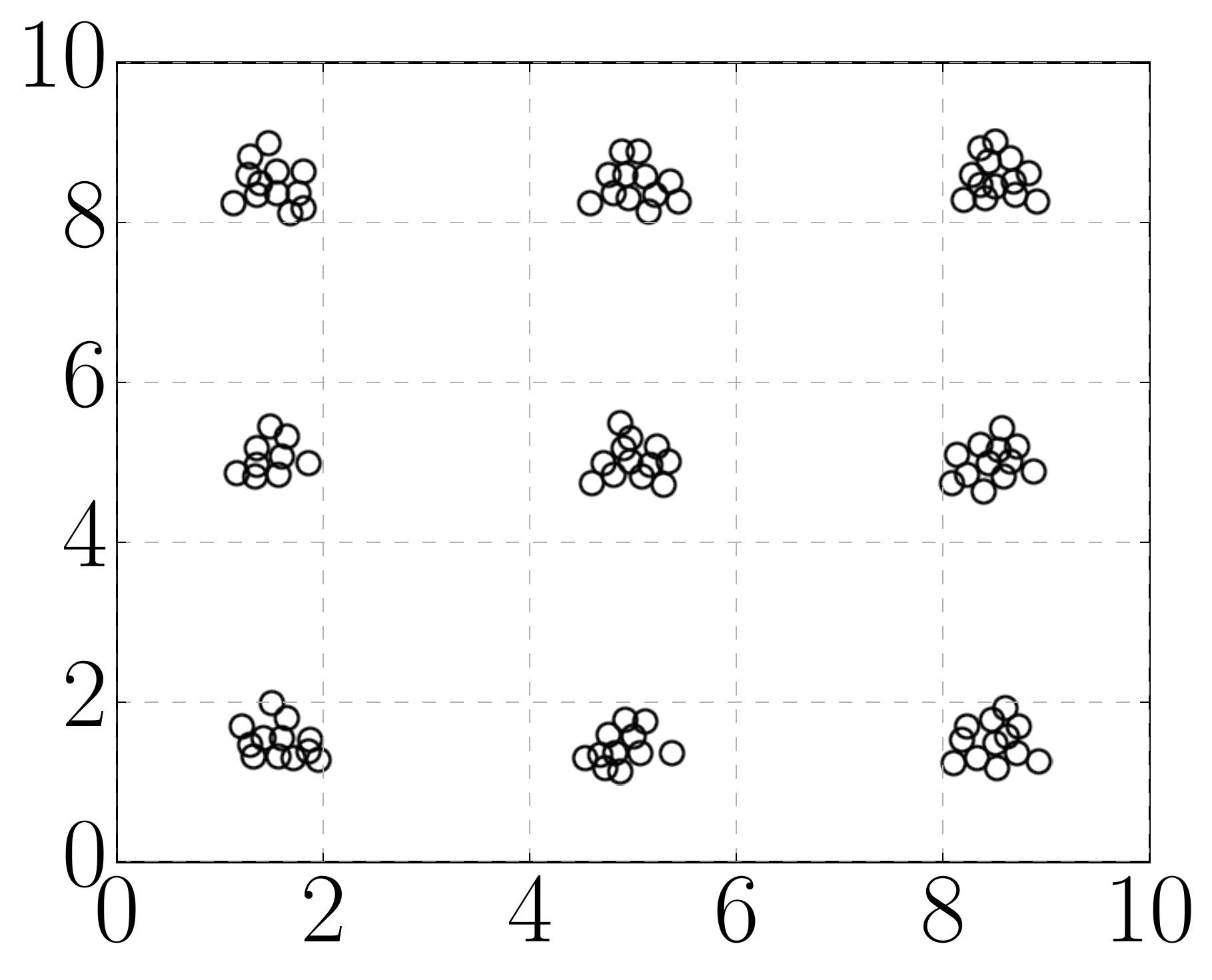}}
\caption{
\small
Distribution of non-dominated solutions found by each method in the solution space on the three-objective Polygon problem.
The x and y axes represent $x_1$ and $x_2$, respectively.
}
\label{fig:3polygon_var}
  \end{center}
\end{figure}

\subsubsection{Distribution of solutions}


Figs. \ref{fig:3polygon_obj} and \ref{fig:3polygon_var} show the distribution of non-dominated solutions found by eMOEA/D-ADA, NSGA-III-ADA, RVEA-ADA, and their original versions on Polygon with $M=3$ in the objective and solution spaces, respectively.
These figures show results of a single run with a median IGD$^+$ and IGDX values, respectively.
For the ADA-based algorithms, non-dominated solutions in $\vector{A}^{\rm primary}$ and $\vector{A}^{\rm tertiary}$ are shown in Figs. \ref{fig:3polygon_obj} and \ref{fig:3polygon_var}, respectively.
Figs. S.1 and S.2 in the supplementary file show the results of the other ADA-based algorithms, but the results are similar to Figs. \ref{fig:3polygon_obj} and \ref{fig:3polygon_var}.
It was shown in \cite{DebJ14} that some decomposition-based EMOAs did not work well on problems with convex Pareto fronts.
It was also presented in \cite{IshibuchiSMN16} that some decomposition-based EMOAs performed poorly on problems with inverted-triangular Pareto fronts.
Since Polygon has a convex and an inverted triangular Pareto front, it is difficult for decomposition-based EMOAs to find good solutions.



For the above-mentioned reasons, no method in Fig. \ref{fig:3polygon_obj} can approximate the Pareto front of Polygon well.
Nevertheless, some ADA-based algorithms (e.g., eMOEA/D-ADA and RVEA-ADA) find better distributed objective vectors than their original EMOAs.
This unintended effect of ADA may be caused by the diversity of the population in the solution space \cite{UlrichBZ10}.
Although we do not claim that ADA can improve the performance of EMOAs for MOPs in addition to MMOPs, the solution space diversity maintenance might help the original EMOAs to handle problems with irregular Pareto fronts.




In the Polygon problem with $M=3$ and $O=9$, equivalent Pareto solution sets are inside of the nine regular triangles in the solution space.
Figs. \ref{fig:3polygon_var} (a)--(c) show that the original EMOAs cannot locate all equivalent Pareto solution sets well.
In contrast, Figs. \ref{fig:3polygon_var} (d)--(f) demonstrate that their ADA versions can locate all nine equivalent Pareto solution sets.
Results on other test problems are similar to Fig. \ref{fig:3polygon_var}.

\begin{table}[t!]
\renewcommand{\arraystretch}{1}
\centering
\caption{\small Results of the nine EMMAs on the 15 test problem instances.
Tables (a) and (b) show the APS values of the algorithms for the IGD$^+$ and IGDX indicators, respectively.
The numbers in parentheses are the ranks of the algorithms based on their APS values.
  }
  \label{tab:comp_three_methods}
  \subfloat[IGD$^+$]{
    {\scriptsize
      \scalebox{0.97}[1]{
        \begin{tabular}{|l|c|}                  
          \hline
          MOEA/D-AGR-ADA & 3.7 (7) \\ \hline
          MOEA/D-DU-ADA & 2.8 (5) \\ \hline
          eMOEA/D-ADA & 2.9 (6) \\ \hline
          NSGA-III-ADA & 1.7 (2) \\ \hline
          $\theta$-DEA-ADA & 2.7 (4) \\ \hline
          RVEA-ADA & 5.9 (9) \\ \hline
          TriMOEA-TA\&R & 2.4 (3) \\ \hline
          MO\_Ring\_PSO\_SCD & 4.2 (8) \\ \hline
          Omni-optimizer & \cellcolor{c1}1.6 (1) \\ \hline
        \end{tabular}
      }
    }
  }
  \subfloat[IGDX]{
    {\scriptsize
      \scalebox{0.97}[1]{
        \begin{tabular}{|l|c|}                  
          \hline
          MOEA/D-AGR-ADA & 2.1 (3) \\ \hline
          MOEA/D-DU-ADA & 2.1 (4) \\ \hline
          eMOEA/D-ADA & 2.3 (5) \\ \hline
          NSGA-III-ADA & \cellcolor{c1}0.8 (1) \\ \hline
          $\theta$-DEA-ADA & 1.0 (2) \\ \hline
          RVEA-ADA & 4.1 (6) \\ \hline
          TriMOEA-TA\&R & 7.7 (9) \\ \hline
          MO\_Ring\_PSO\_SCD & 4.3 (7) \\ \hline
          Omni-optimizer & 5.1 (8) \\ \hline
        \end{tabular}
      }
    }
  }

\end{table}

\subsection{Comparison with other EMMAs}
\label{sec:comparison_three_methods}


In Subsection \ref{sec:comparison_ada}, we demonstrated that ADA can improve the performance of the six EMOAs for MMOPs.
Here, we compare the six ADA-based algorithms to the following three EMMAs: Omni-optimizer \cite{DebT08}, MO\_Ring\_PSO\_SCD \cite{YueQL17}, and TriMOEA-TA\&R \cite{LiuYG18}.
Omni-optimizer is the most representative EMMA.
MO\_Ring\_PSO\_SCD and TriMOEA-TA\&R are recently proposed methods.
Default parameter settings were used for the three methods.

Table \ref{tab:comp_three_methods} shows the APS values of the nine EMMAs on the 15 test problem instances.
Table S.2 in the supplementary file shows detailed results.
TriMOEA-TA\&R incorrectly recognizes that some problems have distance-related variables (e.g., SYM-PART1).
In such a case, TriMOEA-TA\&R generates a set of additional solutions $\vector{A}^{\rm add}$ by recombining solutions in the diversity archive and the distance-related variables at the end of the search.
Since $|\vector{A}^{\rm add}| > N$, methods of selecting $N$ solutions from $\vector{A}^{\rm add}$ are needed.
We use the two post-processing methods in ADA. 
In the same manner as in ADA, $\vector{A}^{\rm primary}$ and $\vector{A}^{\rm tertiary}$ are used for the IGD$^+$ and IGDX calculations, respectively.
We use the selection method in NSGA-III-ADA to obtain $\vector{A}^{\rm primary}$ from $\vector{A}^{\rm add}$.

Table \ref{tab:comp_three_methods} (a) shows that Omni-optimizer performs the best regarding IGD$^+$.
The good performance of Omni-optimizer for MOPs is consistent with the results presented in \cite{YueQL17,TanabeI18}.
NSGA-III-ADA shows the best performance regarding IGD$^+$ among the six ADA-based algorithms.
Table \ref{tab:comp_three_methods} (b) shows that the six ADA-based algorithms perform better than the three other algorithms regarding IGDX.
TriMOEA-TA\&R performs the worst regarding IGDX.
This is because the 15 problem instances have no distance-related variables.
In summary, the results indicate that the six ADA-based algorithms have better performance than the state-of-the-art algorithms for MMOPs.





\begin{figure}[t]
  \newcommand{\widthvar}{0.37}
  %
  \begin{center}
\includegraphics[width=\widthvar\textwidth]{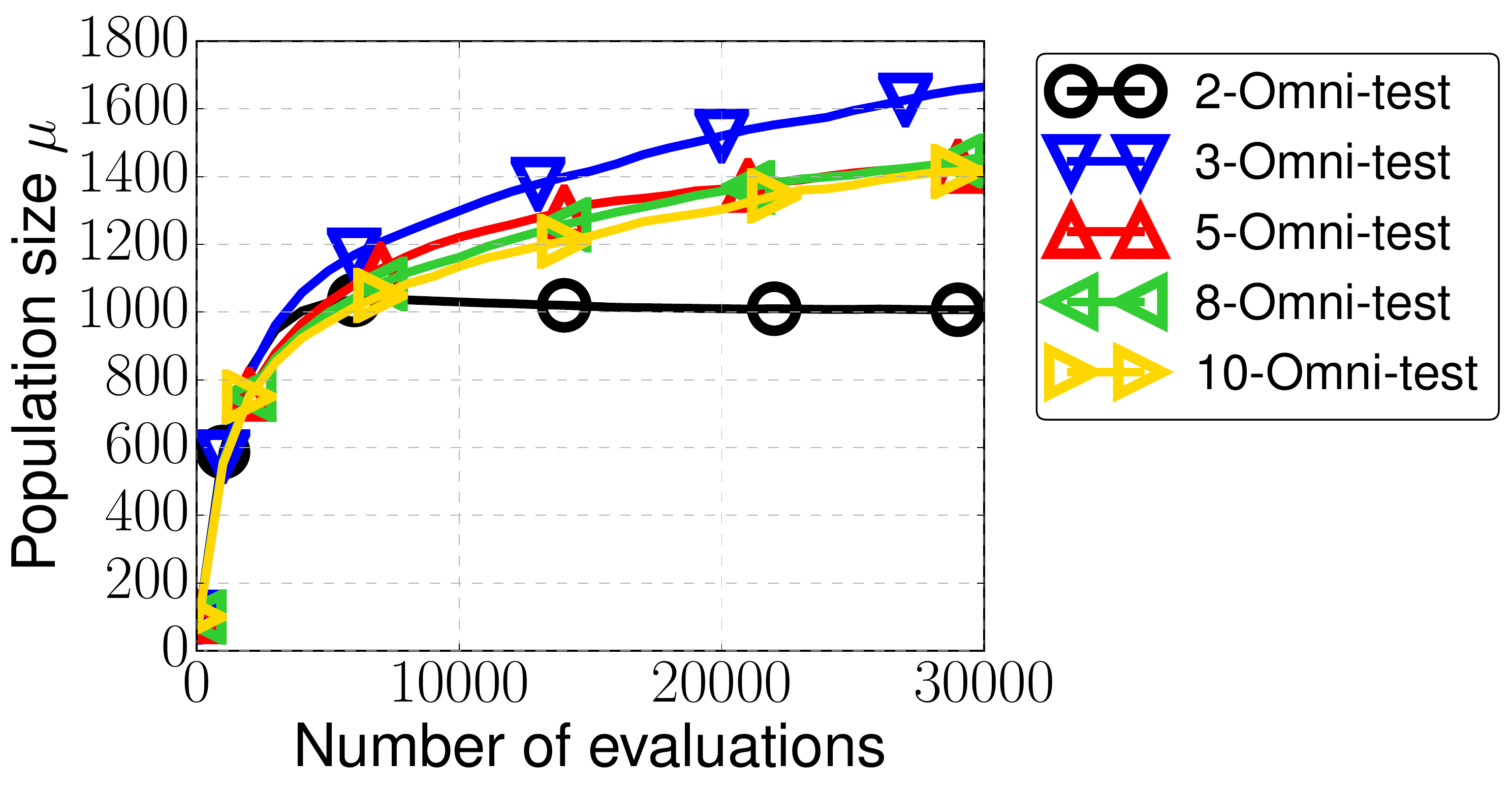}
    \caption{
\small
%
Evolution of $\mu$ of NSGA-III-ADA on Omni-test with $D \in \{2, 3, 5, 8, 10\}$.
The mean $\mu$ values over 31 runs are reported.
}
\label{fig:nsgaiiiada_3popygon_popsize_omni-test}
  \end{center}
%
  \begin{center}
\includegraphics[width=0.3\textwidth]{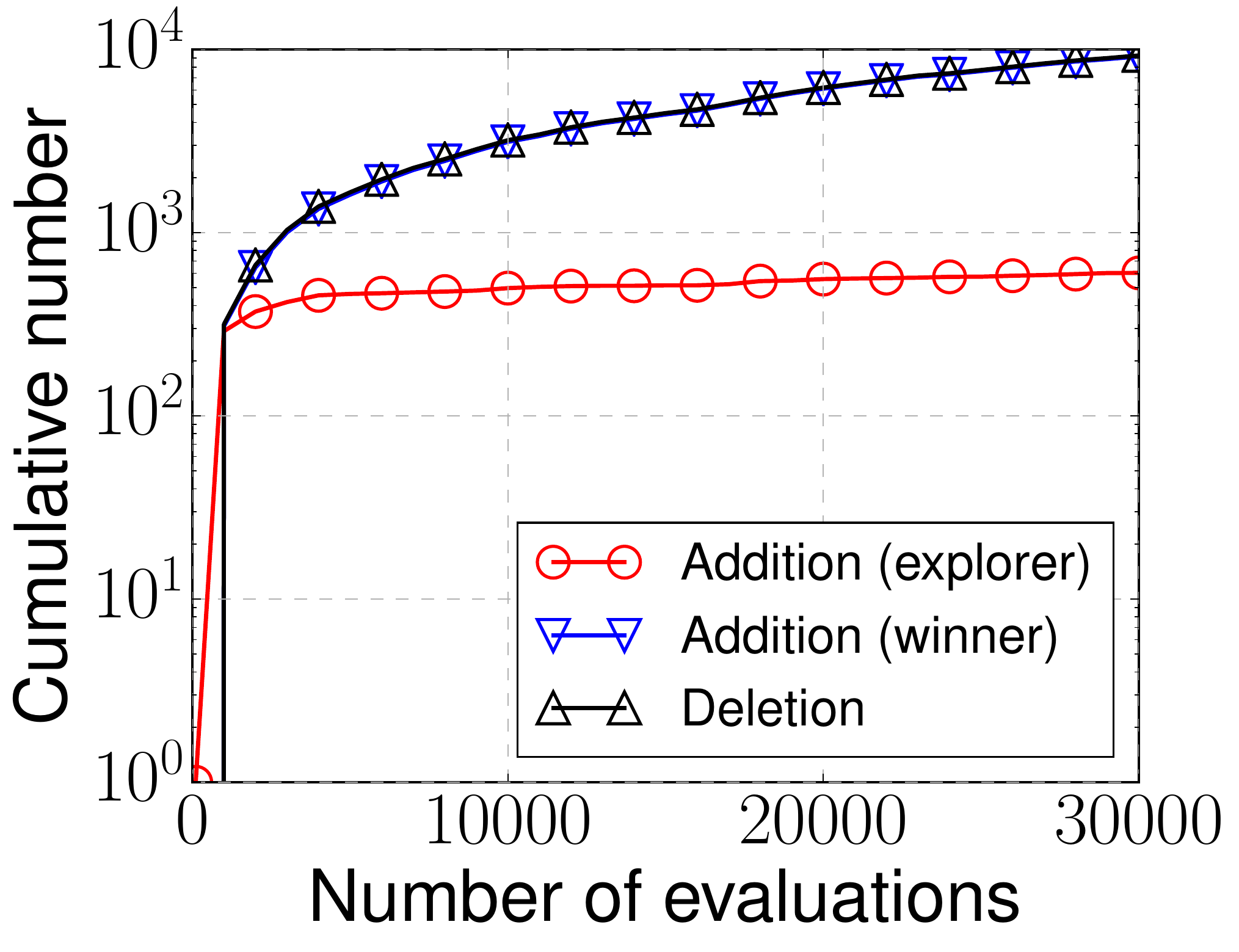}
    \caption{
\small
Cumulative number of the activations of the addition operation based on the two criteria ($b^{\rm explorer}$ and $b^{\rm winner}$) and the deletion operation in NSGA-III-ADA on 3-Polygon ($M=3$).
Results of a single run with a median IGDX value among 31 runs are shown.
}
\label{fig:nsgaiiiada_cumnum}
  \end{center}
\end{figure}

\subsection{Adaptive population sizing in ADA}
\label{sec:results_popsize}

In ADA, the number of individuals assigned to each subproblem is adaptively adjusted by the deletion and addition operations.
Thus, the population size $\mu$ is not constant.
Here, we discuss the adaptive population sizing in ADA.





Fig. \ref{fig:nsgaiiiada_3popygon_popsize_omni-test} shows the change of $\mu$ of NSGA-III-ADA on Omni-test with $D \in \{2, 3, 5, 8, 10\}$.
Results of the other ADA versions are similar to Fig. \ref{fig:nsgaiiiada_3popygon_popsize_omni-test}.
The $\mu$ value for $D=3$ is always larger than that for $D=2$.
In contrast, the $\mu$ value decreases as the $D$ value increases from $D = 3$.
This may be because problems with a large $D$ value are generally difficult for EMMAs.
Since the search does not proceed well for $D \in \{5, 8, 10\}$, the $\mu$ value does not significantly increase.
The difficulty of finding multiple equivalent Pareto sets has been reported for the case of a large $D$ value \cite{IshibuchiPS19}.

In addition, the trajectory of $\mu$ is problem-dependent.
Fig. S.3 in the supplementary file shows the change of $\mu$ on other problems.
While the $\mu$ value is approximately $640$ on Two-On-One and 3-Polygon at the end of the search, it is approximately $960$--$1\,090$ on other problems.
Fig. S.3 shows that the $\mu$ value sharply increases in early iterations.
After some evaluations, the $\mu$ value is stable.

Fig. \ref{fig:nsgaiiiada_cumnum} shows the cumulative numbers of the activations of the addition operation based on the two criteria ($b^{\rm explorer}$ and $b^{\rm winner}$ in Algorithm \ref{alg:ada}) and the deletion operation in NSGA-III-ADA on 3-Polygon.
On the one hand, the number of activations of the addition operation based on $b^{\rm winner}$ is almost the same as that of the deletion operation throughout the evolution in Fig. \ref{fig:nsgaiiiada_cumnum}.
This competitive behavior of the two operations leads to the stable $\mu$ value in Fig. S.3.
On the other hand, the addition operation based on $b^{\rm explorer}$ is frequently performed only in an early stage of evolution in Fig. \ref{fig:nsgaiiiada_cumnum}.
Thus, the sharp increase of the $\mu$ value in Fig. S.3 is due to the $b^{\rm explorer}$-based addition operation.
Since only one individual is assigned to each subproblem at the beginning of the search, the $b^{\rm explorer}$-based addition operation is often activated.


In summary, the $\mu$ value is adaptively adjusted by the cooperative behavior of the deletion and addition operations in ADA.
In the addition operation, the roles of the two criteria $b^{\rm explorer}$ and $b^{\rm winner}$ differ from each other.
Thus, it is important for ADA to use both $b^{\rm explorer}$ and $b^{\rm winner}$.

\subsection{Impact of $L$ on the six ADA-based algorithms}
\label{sec:impact_l}



We investigate the influence of the $L$ value on the performance of the ADA-based algorithms.
Table \ref{tab:invesigation_L} shows the APS values of the six ADA-based algorithms with six $L$ values on the 15 test problem instances.
Tables S.3--S.8 in the supplementary file show detailed results.

Table \ref{tab:invesigation_L} (a) shows that the six ADA-based algorithms with $L = \lfloor 0.05 \mu \rfloor$ have the worst performance regarding IGD$^+$.
Too small $L$ values degrade the performance of the ADA-based algorithms regarding IGD$^+$.
Table \ref{tab:invesigation_L} (b) shows that $L =\lfloor 0.1 \mu \rfloor$ is most suitable for all ADA-based algorithms (except for RVEA-ADA) in terms of IGDX.
Too large $L$ values degrade the performance of the ADA-based algorithms regarding IGDX.
In summary, $L =\lfloor 0.1 \mu \rfloor$ are suitable for most ADA-based algorithms for MMOPs.


We also investigate the influence of $L$ on the evolution of $\mu$.
Figs. S.4 and S.5 in the supplementary file show the change of $\mu$ of NSGA-III-ADA with various $L$ values on Omni-test with various $D$ values and other problems, respectively.
Results of other ADA versions are similar to the results of NSGA-III-ADA.
Figs. S.4 and S.5 indicate that the $\mu$ value decreases as the $L$ value increases.
The larger the $L$ value is, the more individuals can be neighbors of the child $\vector{u}$.
As a result, the niching mechanism in ADA deteriorates.
These observations are consistent with the above-mentioned results of IGDX.

\begin{table}[t!]
\renewcommand{\arraystretch}{1}
\centering
\caption{\small Results of the six ADA-based algorithms with $L \in \{\lfloor 0.05 \mu \rfloor, \lfloor 0.1 \mu \rfloor, \lfloor 0.2 \mu \rfloor, \lfloor 0.3 \mu \rfloor, \lfloor 0.4 \mu \rfloor, \lfloor 0.5 \mu \rfloor \}$ on the 15 test problem instances.
Tables (a) and (b) show the APS values of the algorithms for the IGD$^+$ and IGDX indicators, respectively.
AGR and DU stand for MOEA/D-AGR and MOEA/D-DU, respectively.
  }
  \label{tab:invesigation_L}
  \subfloat[IGD$^+$]{
    {\scriptsize
      \scalebox{0.89}[1]{
      \begin{tabular}{|l|c|c|c|c|c|c|}                  
        \hline
        & $\lfloor 0.05\mu \rfloor$ & $\lfloor 0.1\mu \rfloor$ & $\lfloor 0.2\mu \rfloor$ & $\lfloor 0.3\mu \rfloor$ & $\lfloor 0.4\mu \rfloor$ & $\lfloor 0.5\mu \rfloor$\\
      \hline
        AGR-ADA & 1.6 (6)  & 0.9 (3)  & \cellcolor{c1}0.6 (1)  & 0.7 (2)  & 1.0 (4)  & 1.3 (5) \\ \hline
        DU-ADA & 1.6 (6)  & 0.7 (2)  & \cellcolor{c1}0.5 (1)  & 0.7 (3)  & 1.0 (4)  & 1.3 (5) \\ \hline
        eMOEA/D-ADA  & 1.7 (6)  & 0.9 (3)  & \cellcolor{c1}0.5 (1)  & 0.5 (2)  & 0.9 (4)  & 1.1 (5) \\ \hline
        NSGA-III-ADA & 2.6 (6)  & 1.5 (5)  & 0.7 (4)  & \cellcolor{c1}0.1 (1) & \cellcolor{c1}0.1 (1)  & 0.2 (3) \\ \hline
        $\theta$-DEA-ADA & 2.2 (6)  & 1.3 (5)  & 0.4 (4)  & 0.2 (3)  & \cellcolor{c1}0.0 (1)  & \cellcolor{c1}0.0 (1) \\ \hline
        RVEA-ADA & 1.9 (6)  & \cellcolor{c1}0.5 (1)  & 1.1 (3)  & 0.9 (2)  & 1.3 (4)  & 1.9 (5) \\ \hline
      \end{tabular}
    }
  }
     }
  \\
  \subfloat[IGDX]{
    {\scriptsize
      \scalebox{0.89}[1]{        
      \begin{tabular}{|l|c|c|c|c|c|c|}            
      \hline
        & $\lfloor 0.05\mu \rfloor$ & $\lfloor 0.1\mu \rfloor$ & $\lfloor 0.2\mu \rfloor$ & $\lfloor 0.3\mu \rfloor$ & $\lfloor 0.4\mu \rfloor$ & $\lfloor 0.5\mu \rfloor$\\
      \hline
        AGR-ADA & 0.8 (2)  & \cellcolor{c1}0.1 (1)  & 0.8 (3)  & 2.6 (4)  & 2.9 (5)  & 3.2 (6) \\ \hline
        DU-ADA & 0.8 (2)  & \cellcolor{c1}0.3 (1)  & 1.0 (3)  & 2.5 (4)  & 2.9 (5)  & 3.5 (6) \\ \hline
        eMOEA/D-ADA & 0.9 (2)  & \cellcolor{c1}0.3 (1)  & 1.1 (3)  & 2.2 (4)  & 2.5 (5)  & 3.2 (6) \\ \hline
        NSGA-III-ADA & 1.0 (3)  & \cellcolor{c1}0.1 (1)  & 0.5 (2)  & 2.2 (4)  & 2.7 (5)  & 3.5 (6) \\ \hline
        $\theta$-DEA-ADA & 1.1 (3)  & \cellcolor{c1}0.2 (1)  & 0.9 (2)  & 2.1 (4)  & 2.7 (5)  & 3.2 (6) \\ \hline
        RVEA-ADA & \cellcolor{c1}0.5 (1)  & 0.6 (2)  & 1.6 (3)  & 2.7 (4)  & 2.9 (5)  & 3.1 (6) \\ \hline
      \end{tabular}
      }
    }
  }
%
\end{table}



\subsection{Results on test problems with distance-related variables}
\label{sec:hps_mmmop}



This subsection shows results on the HPS and MMMOP test problems.
Although the six HPS problem instances (HPS1, ..., HPS6) are proposed in \cite{ZhangSA17}, we use only HPS2.
This is because all the other five problem instances are variants of HPS2 and also because the details of only HPS2 are provided in \cite{ZhangSA17}.
$M$, $D$, and $O$ in HPS2 are as follows: $M=2$, $D=7$, and $O=4$.
The 20 MMMOP problem instances (MMMOP1A, ..., MMMOP6D) are proposed in \cite{LiuYG18}.
We also use those 20 instances.
$M$, $D$, and $O$ in MMMOP are as follows: $M \in \{2, 3\}$, $D \in \{2, ..., 7\}$, and $O \in \{2, ..., 9\}$.


Table S.9 in the supplementary file shows results of the six EMOAs and their ADA versions on the 21 problem instances.
%
The results in Table S.9 are similar to the results in Table \ref{tab:comparison}.
Since the solution space diversity maintenance mechanism in ADA is not useful to handle distance-related variables, the ADA-based algorithms are likely to perform poorly on the problems with distance-related variables.
However, the results show that the ADA-based algorithms perform better than their original EMOAs (except for NSGA-III-ADA).

Table \ref{tab:comp_three_methods_on_problems_distance} shows the APS values of the nine EMMAs on the 21 problem instances.
Table S.10 in the supplementary file shows detailed results.
Since TriMOEA-TA\&R can explicitly exploit distance-related variables for the search similar to MOEA/DVA \cite{MaLQWLJYG16}, its performance is improved on the 21 problem instances.
Table \ref{tab:comp_three_methods_on_problems_distance} (a) shows that TriMOEA-TA\&R performs the best regarding IGD$^+$.
Table \ref{tab:comp_three_methods_on_problems_distance} (b) shows that TriMOEA-TA\&R is the fifth-ranked algorithm regarding IGDX.
MOEA/D-AGR-ADA, MOEA/D-DU-ADA, eMOEA/D-ADA, and $\theta$-DEA-ADA perform better than the three competitors regarding IGDX.
In summary, the ADA-based algorithms have high performance for MMOPs even on the problems with distance-related variables.

\begin{table}[t!]
\renewcommand{\arraystretch}{1}
\centering
\caption{\small Results of the nine EMMAs on the 21 test problem instances with distance-related variables: APS value of each algorithm for IGD+ in (a) and for IGDX in (b).
  }
  \label{tab:comp_three_methods_on_problems_distance}
  \subfloat[IGD$^+$]{
    {\scriptsize
      \scalebox{0.97}[1]{
        \begin{tabular}{|l|c|}                  
          \hline
          MOEA/D-AGR-ADA & 3.0 (5) \\ \hline
          MOEA/D-DU-ADA & 2.2 (3) \\ \hline
          eMOEA/D-ADA & 1.5 (2) \\ \hline
          NSGA-III-ADA & 4.2 (7) \\ \hline
          $\theta$-DEA-ADA & 3.0 (6) \\ \hline
          RVEA-ADA & 4.4 (8) \\ \hline
          TriMOEA-TA\&R & \cellcolor{c1}1.0 (1) \\ \hline
          MO\_Ring\_PSO\_SCD & 6.8 (9) \\ \hline
          Omni-optimizer & 2.3 (4) \\ \hline
        \end{tabular}
      }
    }
  }
  \subfloat[IGDX]{
    {\scriptsize
      \scalebox{0.97}[1]{
        \begin{tabular}{|l|c|}                  
          \hline
          MOEA/D-AGR-ADA & 2.0 (3) \\ \hline
          MOEA/D-DU-ADA & 1.5 (2) \\ \hline
          eMOEA/D-ADA & \cellcolor{c1}0.9 (1) \\ \hline
          NSGA-III-ADA & 3.0 (6) \\ \hline
          $\theta$-DEA-ADA & 2.2 (4) \\ \hline
          RVEA-ADA & 3.7 (7) \\ \hline
          TriMOEA-TA\&R & 2.7 (5) \\ \hline
          MO\_Ring\_PSO\_SCD & 6.9 (9) \\ \hline
          Omni-optimizer & 4.1 (8) \\ \hline
        \end{tabular}
      }
    }
  }

\end{table}

\subsection{Comparison using an unbounded external archive}
\label{sec:comparison_three_methods_uea}

While the three EMMAs (TriMOEA-TA\&R, MO\_Ring\_PSO\_SCD, and Omni-optimizer) keep the $\mu$ value constant, the ADA-based algorithms can adaptively adjust the $\mu$ value.
One may think that the comparisons in Subsections \ref{sec:comparison_three_methods} and \ref{sec:hps_mmmop} are unfair for this reason.
Here, we compare the ADA-based algorithms to the three EMMAs using an unbounded external archive (UEA) \cite{TanabeIO17}, which stores all non-dominated solutions found during the search process.
The UEA can be incorporated into any algorithms with no changes in their algorithmic behavior.
The three EMMAs with the UEA can maintain all non-dominated solutions found so far.
Note that only the three EMMAs use the UEA in this section.
Since an algorithm with the UEA significantly outperforms its original version (see \cite{TanabeIO17}), such a comparison is unfair for the ADA-based algorithms.







Tables S.11 and S.12 in the supplementary file show the comparisons with the three EMMAs using the UEA on the 15 test problem instances with no distance-related variables and the 21 test problem instances with distance-related variables, respectively.
The performance of the three EMMAs is improved by using the UEA.
However, Table S.11 indicates that all ADA-based algorithms (except for RVEA-ADA) outperform the three EMMAs with the UEA on the 15 test problems in terms of IGDX.
Table S.12 also shows that eMOEA/D-ADA performs the best on the 21 test problems in terms of IGDX.
Thus, the results show that some ADA-based algorithms perform significantly better than the three EMMAs even with the UEA.
Note that the performance of the ADA-based algorithms can be further improved by using the UEA as in the three EMMAs demonstrated here.

\subsection{On the runtime of the ADA-based algorithms}
\label{sec:runtime}


The space complexity of ADA itself is $\mathcal{O}(\mu D)$ if $D>M$.
Otherwise, it is $\mathcal{O}(\mu M)$.
Although the $\mu$ value equals to the maximum number of function evaluations $n^{\rm max}$ in ADA in the worst case, the $\mu$ value is naturally bounded as shown in Subsection \ref{sec:results_popsize}.
Thus, the empirical space complexity is much smaller than $\mathcal{O}(n^{\rm max} D)$ and $\mathcal{O}(n^{\rm max} M)$.


Below, we explain the worst-case time complexity of the ADA-based algorithms.
ADA itself in one iteration requires $\mathcal{O}(\mu D)$, which is due to the procedure of selecting the neighbors of the child $\vector{u}$ in the solution space (line 11 in Algorithm \ref{alg:ada}).
The time complexity of an ADA-based algorithm depends on its original version.
Since the time complexity of NSGA-III is the larger value between $\mathcal{O}(\mu^2 {\rm log}^{M-2} \mu)$ and $\mathcal{O}(\mu^2 M)$ \cite{DebJ14}, that of NSGA-III-ADA is the largest value among $\mathcal{O}(\mu^2 {\rm log}^{M-2} \mu)$, $\mathcal{O}(\mu^2 M)$, and $\mathcal{O}(\mu D)$.
Similarly, the time complexity of $\theta$-DEA-ADA and RVEA-ADA is the larger value between $\mathcal{O}(\mu^2 M)$ \cite{YuanXWY16,ChenJOS16} and $\mathcal{O}(\mu D)$.
The assignment operations in MOEA/D-AGR, MOEA/D-DU, and eMOEA/D in one function evaluation require $\mathcal{O}(\mu M)$ \cite{WangZZGJ16,YuanXWZY16,JiangYWL18}.
Thus, the time complexity of MOEA/D-AGR-ADA, MOEA/D-DU-ADA, and eMOEA/D-ADA in one iteration is the larger value between $\mathcal{O}(\mu M)$ and $\mathcal{O}(\mu D)$.

\begin{figure}[t]
  \newcommand{\widthvar}{0.39}
  \begin{center}
    \includegraphics[width=\widthvar\textwidth]{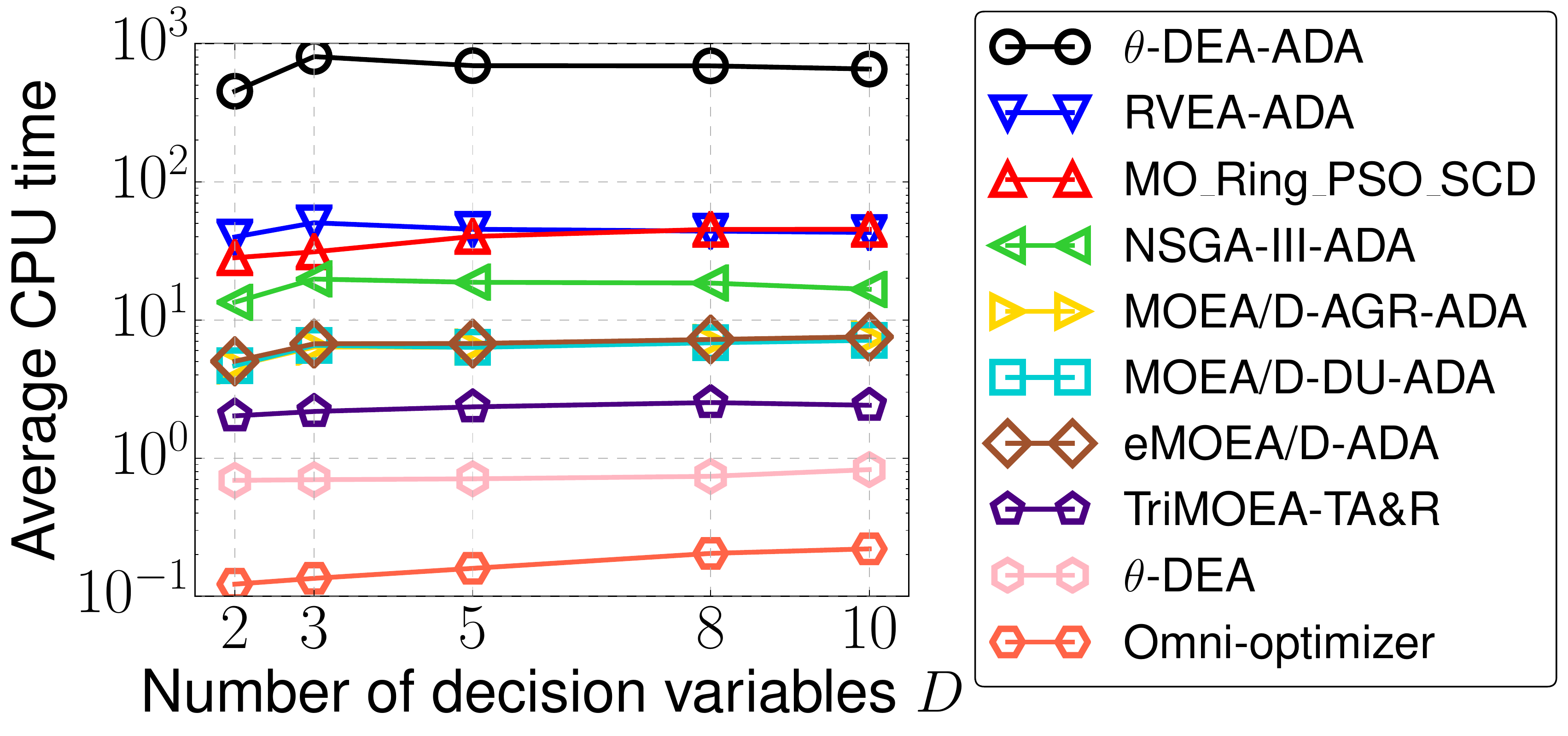}
    \caption{
\small
Average CPU time of the EMMAs over the five Omni-test problems with $D = 2, 3, 5, 8, 10$.
}
\label{fig:runtime}
  \end{center}
\end{figure}

Fig. \ref{fig:runtime} shows the average CPU time of each EMMA over the five Omni-test problems with $D=2, 3, 5, 8, 10$.
We obtained all results using a workstation with Xeon E5-2620 v4/$2.10$GHz and 8GB RAM.
Fig. \ref{fig:runtime} also shows results of $\theta$-DEA.
The six ADA-based algorithms and $\theta$-DEA are implemented in Java, Omni-optimizer is implemented in C, and MO\_Ring\_PSO\_SCD and TriMOEA-TA\&R are implemented in Matlab.
These implementations are based on the available codes of these algorithms.
The use of C and Matlab is due to the original implementations of these algorithms \cite{DebT08,YueQL17,LiuYG18}.
Since different programming languages require different CPU time, Fig. \ref{fig:runtime} provides rough comparison.


As shown in Fig. \ref{fig:runtime}, Omni-optimizer is fastest, followed by $\theta$-DEA and TriMOEA-TA\&R.
The six ADA-based algorithms are much slower than these three algorithms.
This is because the $\mu$ value of the ADA-based algorithms is always larger than that of the other algorithms (see Subsection \ref{sec:results_popsize}).
Since some operations require $\mathcal{O}(\mu^2 M)$ and $\mathcal{O}(\mu D)$ as discussed above, a larger $\mu$ value makes an algorithm slower.
Although MO\_Ring\_PSO\_SCD and RVEA-ADA perform similarly in Fig. \ref{fig:runtime}, we believe that MO\_Ring\_PSO\_SCD with an optimized source code can perform much faster than the ADA-based algorithms.
Clearly, $\theta$-DEA-ADA is slowest.
This is because $\theta$-DEA-ADA needs to normalize objective vectors for every iteration in a computationally expensive manner.
Since the original $\theta$-DEA is a generational EMOA, it can reduce the number of normalizations.
In fact, $\theta$-DEA is much faster than $\theta$-DEA-ADA.
Note that $\theta$-DEA-ADA can be speeded up by using a simple normalization method (e.g., Subsection \ref{sec:ada_normalization}).

As demonstrated here, the slow speed of the ADA-algorithms is their disadvantage.
Speeding up the ADA-algorithms is an avenue for future work.
However, this does not mean that ADA is impractical.
Some real-world problems require a long computation time to evaluate a solution, e.g., by the execution of an expensive computer simulation \cite{ChughSHM19}.
In this case, the time of function evaluations dominates that of the other parts of optimization algorithms.
For this reason, in the EMO community, the comparison is generally based on the number of function evaluations, rather than the CPU time.

\section{Conclusion}
\label{sec:conclusion}

We proposed the ADA framework to improve the performance of decomposition-based EMOAs for MMOPs.
ADA adaptively assigns one or more individuals to each subproblem to locate multiple equivalent Pareto optimal solutions.
The population size $\mu$ is automatically adjusted during the search process.
As presented in Subsection \ref{sec:decision_making}, the effective decision making can be performed using multiple equivalent solutions found by ADA.
We incorporated ADA into the six decomposition-based EMOAs.
We examined the performance of those six ADA-based algorithms on the 15 test problem instances and the 21 test problem instances with distance-related variables.
The results show that ADA can improve the performance of the original EMOAs for MMOPs.
We also analyzed the adaptive behavior of ADA.




Computational overhead in the ADA-based algorithms is their disadvantage (see Subsection \ref{sec:runtime}).
Thus, the unbound population size of ADA is a ``double-edged sword''.
However, as discussed in Subsection \ref{sec:runtime}, the computational overhead in ADA is practically acceptable in real-world applications where solution evaluations need dominant computation time in the whole optimization process.

For a fair comparison, we compared the ADA-based algorithms to the three EMMAs using the UEA in Subsection \ref{sec:comparison_three_methods_uea}.
However, the UEA is inherently a post-processing method since an algorithm cannot take advantage of non-dominated solutions in the UEA during the search process as in the ADA-based algorithms.
For this reason, the comparison with the three EMMAs using the UEA is not totally fair.
A general adaptive population sizing framework which can be combined with any EMMA is needed for a totally fair comparison.




Although we focused on Type1-MMOPs, we believe that ADA is applicable to Type2-MMOPs using a relaxed dominance relation or a relaxed equivalent relation.
An analysis of ADA on Type2-MMOPs is an avenue for future work.
%
%
Other post-processing methods that handle the diversity in both the objective and solution spaces may be better than the two post-processing methods in ADA.
Further analysis of post-processing methods is a future research topic.

 \section*{Acknowledgement}


This work was supported by National Natural Science Foundation of China (Grant No. 61876075), the Program for Guangdong Introducing Innovative and Enterpreneurial Teams (Grant No. 2017ZT07X386), Shenzhen Peacock Plan (Grant No. KQTD2016112514355531), the Science and Technology Innovation Committee Foundation of Shenzhen (Grant No. ZDSYS201703031748284), and the Program for University Key Laboratory of Guangdong Province (Grant No. 2017KSYS008).







\ifCLASSOPTIONcaptionsoff
  \newpage
\fi



%



\bibliography{reference}
\bibliographystyle{IEEEtran}

\begin{IEEEbiography}[{\includegraphics[width=1in,height=1.25in,keepaspectratio]{graph/tanabe.pdf}}]{Ryoji Tanabe}
  is a Research Assistant Professor with Department of Computer Science and Engineering, Southern University of Science and Technology, China.
He was a Post-Doctoral Researcher with ISAS/JAXA, Japan, from 2016 to 2017.
He received his Ph.D. in Science from The University of Tokyo, Japan, in 2016.
His research interests include stochastic single- and multi-objective optimization algorithms, parameter control in evolutionary algorithms, and automatic algorithm configuration.
  \end{IEEEbiography}


\begin{IEEEbiography}[{\includegraphics[width=1in,height=1.25in,keepaspectratio]{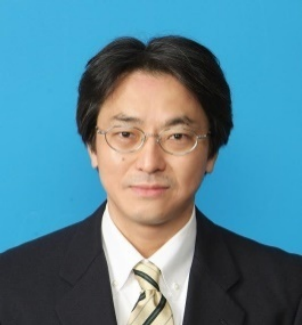}}]{Hisao Ishibuchi}  (M'93-SM'10-F'14)
  received the B.S. and M.S. degrees in precision mechanics from Kyoto University, Kyoto, Japan, in 1985 and 1987, respectively, and the Ph.D. degree in computer science from Osaka Prefecture University, Sakai, Osaka, Japan, in 1992. He was with Osaka Prefecture University in 1987-2017. Since 2017, he is a Chair Professor at Southern University of Science and Technology, China. His research interests include fuzzy rule-based classifiers, evolutionary multi-objective and many-objective optimization, memetic algorithms, and evolutionary games.

Dr. Ishibuchi was the IEEE Computational Intelligence Society (CIS) Vice-President for Technical Activities (2010-2013), an AdCom member of the IEEE CIS (2014-2019), and the Editor-in-Chief of the IEEE COMPUTATIONAL INTELLIGENCE MAGAZINE (2014-2019).
\end{IEEEbiography}

%








\end{document}